\documentclass[a4,journal]{IEEEtran}
\usepackage{amsmath,amsfonts}
\usepackage{algorithmic}
\usepackage{algorithm}
\usepackage{array}
\usepackage[caption=false,font=normalsize,labelfont=sf,textfont=sf]{subfig}
\usepackage{textcomp}
\usepackage{stfloats}
\usepackage{url}
\usepackage{verbatim}
\usepackage{graphicx}
\usepackage{cite}
\usepackage{makecell}
\usepackage{hyperref}
\hypersetup{
colorlinks=true,
linkcolor=blue,
filecolor=magenta,      
citecolor=blue,
urlcolor=green!60!black,
}
\usepackage{tabularx}
\usepackage{colortbl}
\usepackage{float}

\usepackage{booktabs}      % 支持 \toprule、\midrule、\cmidrule 等格式
\usepackage{multirow}       
\usepackage{caption}       % 自定义 caption 样式（可选）
\usepackage[table]{xcolor} % 支持表格配色（如 \rowcolor，用于增强可读性）

\usepackage[most]{tcolorbox}
\tcbuselibrary{skins}
\tcbuselibrary{breakable}
\newtcbox{\textbox}[1][red]
  {on line, arc = 0pt, outer arc = 0pt,
    colback = #1!10!white, colframe = #1!50!black,
    boxsep = 0pt, left = 1pt, right = 1pt, top = 2pt, bottom = 2pt,
    fontupper=\ttfamily\bfseries\upshape,
    boxrule = 0pt, bottomrule = 1pt, toprule = 1pt}
    
\usepackage{environ}  % 用来存 abstract 和 keywords

\NewEnviron{myabstract}{%
  \global\def\savedabstract{\BODY}%
}
\newcommand{\savedabstract}{Dexterous grasp generation aims to produce grasp poses that align with task requirements and human‑interpretable grasp semantics. However, achieving semantically controllable dexterous grasp synthesis remains highly challenging due to the lack of unified modeling of multiple semantic dimensions, including grasp taxonomy, contact semantics, and functional affordance. To address these limitations, we present OmniDexVLG, a multimodal, semantics‑aware grasp generation framework capable of producing structurally diverse and semantically coherent dexterous grasps under joint language and visual guidance.
Our approach begins with OmniDexDataGen, a semantic‑rich dexterous grasp dataset generation pipeline that integrates grasp‑taxonomy–guided configuration sampling, functional‑affordance contact point sampling, taxonomy‑aware differential force‑closure grasp sampling, and physics‑based optimization and validation, enabling systematic coverage of diverse grasp types. We further introduce OmniDexReasoner, a multimodal grasp‑type semantic reasoning module that leverages multi‑agent collaboration, retrieval‑augmented generation (RAG), and chain‑of‑thought (CoT) reasoning to infer grasp‑related semantics and generate high‑quality annotations that align language instructions with task‑specific grasp intent. Building upon these components, we develop a unified Vision-Language-Grasping (VLG) generation model that explicitly incorporates grasp taxonomy, contact structure, and functional affordance semantics, enabling fine‑grained control over grasp synthesis from natural language instructions.
Extensive experiments in simulation and real‑world object grasping and ablation studies demonstrate that our method substantially outperforms state‑of‑the‑art approaches in terms of grasp diversity, contact semantic diversity, functional affordance diversity, and semantic consistency. These results highlight the critical role of multi‑dimensional semantic modeling, including grasp taxonomy, contact semantics, and affordance reasoning, in advancing dexterous grasp generation. More details are available on our project website
~\url{https://sites.google.com/view/omnidexvlg}.
}  % 默认空

\NewEnviron{mykeywords}{%
  \global\def\savedkeywords{\BODY}%
}
\newcommand{\savedkeywords}{
LLM/VLM, Multi-Fingered Robotic Hand, Grasp Taxonomy, Generation Model.
}  % 默认空

\makeatletter

\newcommand{\mycolortitlebox}{%
  \begin{tcolorbox}[
    enhanced,
    breakable,
    colback=green!3,           % 背景颜色
    colframe=green!60!black,   % 边框颜色
    boxrule=0.9pt,
    arc=6pt,
    arc=10pt,
    left=6mm,right=6mm,top=6mm,bottom=6mm
  ]

    \begin{center}
      {\LARGE\bfseries \@title \par}
      \vskip 0.8em
      {\large \@author \par}
    \end{center}

    \vskip 1em

    {\bfseries Abstract — } \savedabstract\par

    \vspace{1em}

    {\bfseries Keywords — } \savedkeywords\par

  \end{tcolorbox}
}

\renewcommand{\maketitle}{%
  \begingroup
    \normalfont

    \setcounter{footnote}{0}%
    \renewcommand{\thefootnote}{\fnsymbol{footnote}}%

    \if@twocolumn
      \twocolumn[
        \begin{@twocolumnfalse}
          \mycolortitlebox
          \vspace{1.0em}
        \end{@twocolumnfalse}
      ]
    \else
      \mycolortitlebox
    \fi

    \setcounter{footnote}{0}%
    \renewcommand{\thefootnote}{\arabic{footnote}}%

    \thispagestyle{empty}
  \endgroup
}

\makeatother

\begin{document}
%%%%%%%%%%%%%%%%%%%%%%%%%%%%%%%%%%%%%%%%%%
% \title{A Sample Article Using IEEEtran.cls\\ for IEEE Journals and Transactions}
% \title{FunctionalDexGraspNet: Multimodal Language Language Model-Guided Grasp Type-Aware Dexterous Multi-Fingered Hand Pose Generation}
% \title{GraspSemNet: Learning Dexterous Grasp Generation from VLM-Guided Grasp Taxonomy, Task, and Region Semantic Understanding}
% \title{VLM-SemDex: Learning Dexterous Grasp Generation from Vision Language Model-Guided Grasp Semantics and Taxonomy}
\title{OmniDexVLG: Learning Dexterous Grasp Generation from Vision Language Model-Guided Grasp Semantics, Taxonomy and Functional Affordance}

% \author{Lei Zhang\textsuperscript{1,2\dag}, Diwen Zheng\textsuperscript{3,2}, Kaixin Bai\textsuperscript{1,2}, Zhenshan Bing\textsuperscript{3}, Zoltan-Csaba Marton\textsuperscript{2}, \\ Zhaopeng Chen\textsuperscript{2}, Alois Christian Knoll\textsuperscript{3}, Jianwei Zhang\textsuperscript{1}
%     \thanks{\dag Corresponding author. lei.zhang-1@studium.uni-hamburg.de}
% \thanks{\textsuperscript{1}MIN-Fakultät Fachbereich Informatik TAMS, University of Hamburg \texttt{\{name.surname\}@studium.uni-hamburg.de}}% <-this % stops a space

% % \author{Anonymous submission,\\
% % Paper ID }
% % 
% \thanks{\textsuperscript{2}Agile Robots SE \texttt{\{name.surname\}@agile-robots.com}}
% \thanks{\textsuperscript{3}Technical University of Munich \texttt{\{name.surname\}@tum.de}}
% }

%  author for new heading
\author{
{ 
Lei Zhang$^{1,2\dagger}$,
Diwen Zheng$^{3,2}$,
Kaixin Bai$^{1,2}$,
Zhenshan Bing$^{3}$,
% Zoltan-Csaba Marton$^2$,\\
Zoltán-Csaba Márton$^{2}$,\\
Zhaopeng Chen$^2$,
Alois Christian Knoll$^{3}$,
Jianwei Zhang$^1$}\\
{\small
$^1$ University of Hamburg \quad
$^2$ Agile Robots SE \quad 
$^3$ Technical University of Munich
}\\[0.1em]
{\small $\dag$ Corresponding Author:~\href{mailto:lei.zhang-1@studium.uni-hamburg.de}
     {\textcolor{green!60!black}{lei.zhang-1@studium.uni-hamburg.de}},~\href{mailto:zhanglei.cn.de@gmail.com}
     {\textcolor{green!60!black}{zhanglei.cn.de@gmail.com}}}
}

% \markboth{IEEE Transactions on Name of Journal}%
\markboth{ }%
{Shell \MakeLowercase{\textit{et al.}}: A Sample Article Using IEEEtran.cls for IEEE Journals}
\maketitle
\section{Introduction}
\label{sec:Intro}

\begin{figure*}[htbp]
  \centering
  \includegraphics[width=1.0\linewidth]{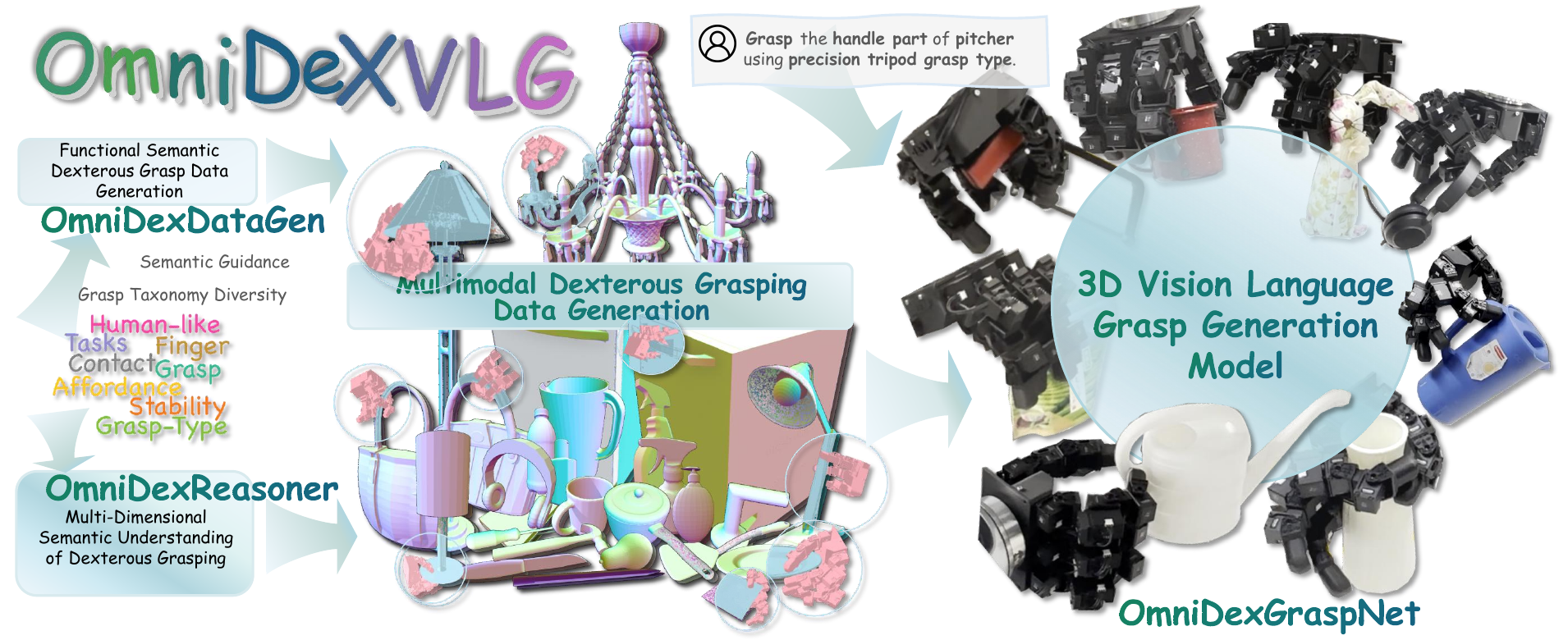}
  \caption{%
  Overview of the OmniDexVLG framework. The framework integrates three core components: OmniDex-DataGen, for functional and grasp taxonomy-aware dexterous grasp dataset generation; OmniDexReasoner, for multi-dimensional semantic understanding using large multimodal models; and OmniDexGraspNet, a 3D vision-language grasp generation model guided by semantic instructions across grasp type, affordance, contact, and finger configuration. }
  \label{fig:heading_photo}
\end{figure*}
Dexterous hands, owing to their high degrees of freedom and complex manipulation capabilities, have demonstrated remarkable potential in fine-grained grasping and coordinated multi-finger tasks. In recent years, advancements in generative models and reinforcement learning have enabled researchers to synthesize reliable dexterous grasp poses from visual observations of objects, substantially improving grasp robustness and generalization across diverse objects and environments~\cite{wang2022dexgraspnet,murali2025graspgen}.

In real-world scenarios, dexterous hands are often required to perform grasping and manipulation tasks with multi-dimensional semantic intent. For example, a task might involve grasping the handle of a kettle using a tripod grasp with the thumb, index, and middle fingers. For humans, such semantic reasoning is naturally achievable through prior knowledge and experience. To model this reasoning process, the concept of grasp taxonomy~\cite{feix2015grasp} has emerged as a structured framework that categorizes grasp strategies based on finger, contact point configurations, opposition types, and force directions, thereby formalizing the diversity of grasp behaviors in both human and robotic hands.

However, existing semantic-guided grasp generation approaches primarily focus on functional grasping, lacking explicit sensitivity to grasp taxonomy and contact semantics~\cite{tang2023graspgpt,tang2025foundationgrasp,li2024semgrasp,huang2025fungrasp,agarwal2023dexterous}. In the context of objects with multiple functional affordance regions, different task goals often require distinct grasp strategies beyond the simplistic alignment with functional parts. Critical semantic elements such as finger flexion, contact configurations, and force application are highly correlated with the intended grasp type, yet current models struggle to utilize such fine-grained cues, resulting in limited controllability and semantic diversity in generated grasp actions.

Most existing dexterous grasp datasets~\cite{wang2022dexgraspnet,zhang2024contactdexnet} are constructed through optimization-based methods that guide hand-object distances while satisfying quality metrics like differential force closure (DFC)~\cite{liu2021synthesizing}. While some recent works have introduced multimodal conditioning to enhance semantic grounding, these models largely focus on part-level semantics or predefined functional goals. They lack the capacity to model and generalize structured grasp semantics, including contact semantics, grasp taxonomy, and affordance-sensitive representations~\cite{wang2022dexgraspnet,zhang2024contactdexnet,murali2025graspgen,zhong2025dexgrasp}. This leads to sparse distributions of grasp types in existing datasets~\cite{wang2022dexgraspnet,zhong2025dexgrasp}. For instance, datasets like DexGraspNet~\cite{wang2022dexgraspnet} are dominated by simplified grasp types such as the pinch grasp, exposing the limitations of current models in generating diverse contact structures and grasp configurations. Although some studies have introduced manually annotated grasp-type labels~\cite{chen2025dexonomy}, how to automatically generate diverse and semantically meaningful grasp types remains an open challenge. 

To address these limitations, we propose a novel grasp data generation framework (OmniDexDataGen) that integrates functional affordance cues, contact pattern priors, and finger configuration semantics. This enables the synthesis of grasp samples that are not only physically stable, but also semantically rich and structurally diverse—providing stronger priors for downstream reasoning and generation tasks.

Furthermore, even in existing datasets with semantic labels, the grasp-related semantics are often coarse and limited to object-level part annotations~\cite{song2025learning} or basic functional descriptors~\cite{zhang2023functionalgrasp}. There remains a lack of a unified semantic understanding model capable of jointly reasoning about grasp taxonomy, contact semantics, and functional affordance. Given the inherent complexity of dexterous hands, semantic reasoning in this context requires modeling the nuanced relationships between hand-object contact topology, force dynamics, and task-driven semantic objectives. The object’s affordance often constrains which grasp types are viable and, in turn, influences the contact structure and pose planning strategy required for successful manipulation. 

Motivated by these challenges, we introduce a multimodal semantic understanding framework based on large multimodal models (LMMs), named OmniDexReasoner, tailored for reasoning over dexterous grasping tasks. Our framework seeks to unify the modeling of grasp type, contact structure, and functional intent under a shared semantic space, enabling fine-grained understanding and generation of natural language grasp instructions for dexterous hands.

To further bridge the semantic gaps in current grasp generation pipelines, especially in representing grasp taxonomy, contact semantics, and affordance, we propose a vision-language grasp generation (VLG) model, OmniDexGraspNet, for semantically grounded grasp generation. By combining multimodal inputs, including language instructions and visual point clouds, with multi-dimensional semantic modeling, our method enables joint reasoning over complex task goals and grasp semantics. This facilitates the generation of grasp poses that are both semantically aligned and physically plausible. Through explicit modeling of structured semantics and controllable generation, our approach not only improves the diversity and realism of generated grasps, but also lays a foundation for more generalized, task-aware robotic manipulation.

Our Contributions are summarized as follows:
\begin{itemize}
\item \textbf{OmniDexDataGen: Functionality-Aware and Contact-Sensitive Dexterous Grasp Dataset Generation Method with Grasp Taxonomy Diversity.} We propose an optimization-based framework for generating dexterous grasp datasets that are sensitive to grasp taxonomy, hand-object contact points, and functional affordance. 
Our approach substantially enhances the dataset's representational richness in terms of contact guidance and grasp semantics. Furthermore, it improves the diversity of grasp types, contact paradigms, and functional affordances.

\item \textbf{OmniDexReasoner: LMM-Based Functional, Contact, and Taxonomy-Aware Dexterous Grasp Understanding Method.} 
We propose an LMM-powered framework for dexterous grasping that captures multi-level semantic cues across three core dimensions: functional affordance, contact semantics, and grasp taxonomy. By incorporating a multi-agent collaboration mechanism, retrieval-augmented generation (RAG) and Chain-of-Thought (CoT) reasoning, the proposed method addresses key limitations in current multimodal understanding approaches for dexterous hands, substantially enhancing the comprehension of complex grasping behaviors. 

\item \textbf{OmniDexGraspNet: Semantic-Aware 3D Vision-Language Grasp Pose Generation Model.} 
We propose a grasp pose generation method that integrates a 3D vision-language model to guide dexterous grasp synthesis using multi-dimensional semantic information and partial object point clouds. The method enables the generation of grasp poses that are sensitive to various semantic dimensions, including functional intent, grasp taxonomy, and contact configuration. It demonstrates strong generalization across objects of different sizes and categories, producing diverse functional grasps with rich semantic and structural variability. Compared to existing grasp generation approaches, our method exhibits superior semantic sensitivity and grasp diversity.

\end{itemize}

%%%%%%%%%%%%%%%%%%%%%%%%%%%%%%%%%%%%%%%%%%
%%%%%%%%%%%%%%%%%%%%%%%%%%%%%%%%%%%%%%%%%%
\section{Related Work}
\label{sec:LitRev}
%%%%%%%%%%%%%%%%%%%%%%%%%%%%%%%%%%%%%%%%%%
\subsection{Semantic-Aware Dexterous Robotic Grasp Generation}

In recent years, dexterous grasp generation has garnered increasing attention with the rise of generative models such as diffusion models~\cite{weng2024dexdiffuser,liu2025ifg,wang2024tooleenet} and variational autoencoders~\cite{zhang2024contactdexnet}, and reinforcement learning~\cite{lum2024dextrah,qi2023general} in the robotics community. Owing to the high DoF and complex manipulation requirements of dexterous hands, grasp synthesis involves rich semantic dimensions, such as grasp types, functional affordance, and contact semantics. Existing approaches have primarily aimed to improve grasp stability and generalization across diverse object categories~\cite{zhong2025dexgrasp,wang2022dexgraspnet}.

To improve semantic grounding, recent works incorporate natural language guidance into grasp generation by leveraging LLMs or VLMs~\cite{li2024semgrasp,li2024multi}. These methods typically target functional grasping, where language instructions are encoded and fused with visual inputs to generate semantically aligned actions.

The complexity of semantic modeling also varies by end-effector type. For two-finger grippers, the semantics are often limited to object-level affordances~\cite{tang2023graspgpt,tang2025foundationgrasp}, whereas dexterous hands require finer modeling of grasp taxonomy, contact semantics, finger configurations, and opposition types. To this end, works~\cite{zhang2023functionalgrasp,wang2025scaleadfg,wu2024cross} exploit hand-object representations as priors for functional grasping generation and transfer.

Moreover, grasp taxonomy and contact semantics have recently received increased attention. For instance, Dexonomy~\cite{chen2025dexonomy} and AnyDexGrasp~\cite{fang2025anydexgrasp} introduce grasp-type encodings to guide grasp synthesis, while ContactDexNet~\cite{zhang2024contactdexnet}, GrainGrasp~\cite{zhao2024graingrasp} and Grasp as You Say~\cite{wei2024grasp} leverage contact maps or linguistic references to inform fine-grained contact configurations.

In summary, while recent advancements have explored various aspects of semantic-aware grasp generation, a unified modeling framework that jointly captures grasp types, contact structures, functional intent, and finger configurations remains lacking. The role of semantic information in guiding dexterous grasp synthesis has yet to be fully explored and systematically leveraged.

\subsection{Dexterous Grasp Taxonomy and Reasoning}

Grasp taxonomy~\cite{feix2015grasp} serves as a fundamental abstraction for describing human hand-object interaction strategies. It provides a structured categorization of grasp types based on contact point locations, involved anatomical links, opposition types, and force directions. 
Classical taxonomies, such as those proposed in~\cite{feix2015grasp}, divide grasps into high-level categories including power, precision, and intermediate grasps, and further into subtypes such as tripod, lateral, and 2-finger pinch. These categories encapsulate nuanced differences in finger articulation, contact configuration, and force application strategies. Notably, recent research continues to identify new grasp types~\cite{liu2021synthesizing}, underscoring both the richness of human and robotic grasp behavior and the inherent complexity of modeling such interactions. This ongoing evolution also reflects the challenge of reasoning over grasp taxonomies, which remains an underexplored area, especially for dexterous hands.

Early works employed CNN-based architectures to infer manipulation semantics, including grasp type and object attributes, directly from visual inputs~\cite{cai2016understanding}. These models built semantic action representations for reasoning about grasp categories and manipulation intent.

Recently, multimodal large models have demonstrated strong capabilities in semantic understanding and spatial comprehension~\cite{ni2025don,jin2024robotgpt,ji2025robobrain,li2024foundation,song2025maniplvm}. 
Language-driven frameworks such as SemGrasp~\cite{li2024semgrasp} and Multi-GraspLLM~\cite{li2024multi} leverage large language models (e.g., GPT-4) to perform grasp type prediction from natural language instructions. However, current vision-language models (VLMs) or large multimodal models (LMMs) struggle with accurate semantic grounding when interpreting hand–object interactions, often exhibiting semantic hallucinations for similar grasp types.

Consequently, there remains a significant gap in designing models that can reason over grasp taxonomy in a physically grounded, task-aware, and semantically aligned manner. Bridging this gap requires novel methods that integrate structured grasp knowledge with multi-modal embeddings, enabling more controllable and diverse dexterous grasp generation.

\begin{figure*}[htbp]
  \centering
  \includegraphics[width=1.0\linewidth]{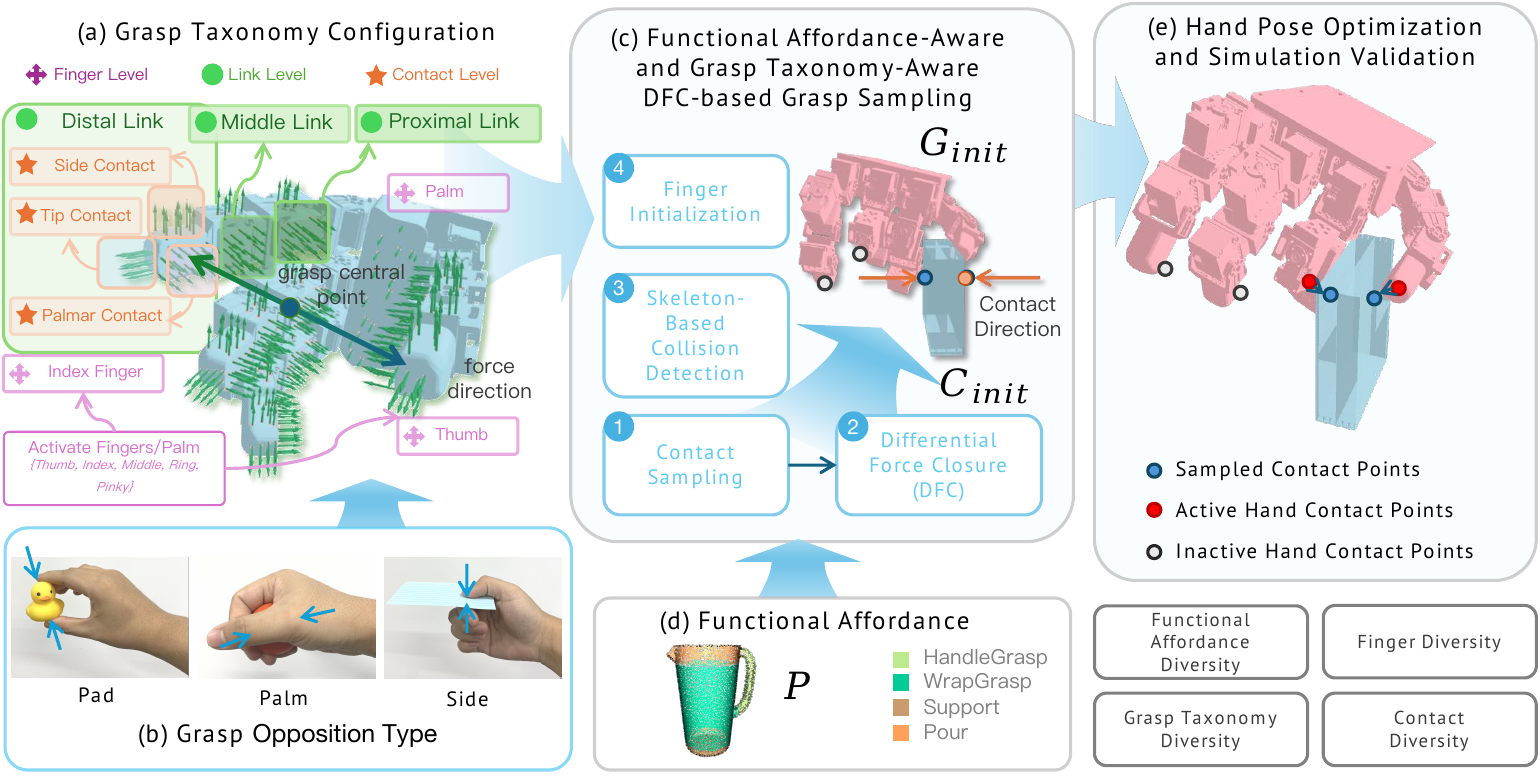}  
  \caption{% 
  OmniDexDataGen: 
  Functional affordance-aware, contact and grasp taxonomy-aware dexterous grasp synthesis. 
  Our grasp synthesis framework introduces a contact-level representation for dexterous hand manipulation. Each grasp is described by a multi-level configuration, including active fingers, involved links, and corresponding contact regions. Based on grasp opposition types, grasp-taxonomy-aware differential force closure sampler (Tax-DFCSampler) computes initial grasp poses and functional affordance contact point sampler (AC-Sampler) conducts contact sampling in object-specific affordance zones. Samples are quantitatively evaluated using differential wrench space metrics. High-quality grasps that pass collision filtering are subjected to hand pose optimization stage, further improving their performance. The finalized grasps are validated through physical simulation to ensure their robustness and applicability.
  }
  \label{fig:grasp_generation_pipeline}
\end{figure*}

\subsection{Dexterous Hand Pose Data Generation}

Existing dexterous grasp datasets~\cite{wang2022dexgraspnet,zhong2025dexgrasp} are predominantly constructed using optimization-based approaches, which aim to generate stable grasp configurations by minimizing hand–object distances or maximizing grasp quality metrics such as Differentiable Force Closure (DFC)~\cite{liu2021synthesizing}. 

In terms of grasp type diversity, datasets such as DexGraspNet~\cite{wang2022dexgraspnet} are heavily biased toward simplified grasp types like the pinch grasp, with limited coverage of more complex or nuanced grasp configurations.  This imbalance stems primarily from the lack of explicit modeling of grasp taxonomy and contact configurations during the data generation phase. 

Other grasp synthesis methods that aim to cover diverse grasp types typically rely on manually predefined parameters or heuristics specific to each grasp category—such as in~\cite{chen2025dexonomy},~\cite{duan2024learning}. However, these approaches lack the capability to automatically generate grasp configurations conditioned on grasp-type-relevant semantics, limiting their adaptability and scalability to unseen tasks or object categories.

While some recent efforts incorporate multimodal inputs~\cite{zhong2025dexgrasp,zhang2024contactdexnet}, the semantic annotations in these datasets remain relatively sparse and coarse. Most labels are restricted to object part names (e.g., handle, tip) or high-level functional descriptions (e.g., pour, hold)~\cite{li2024multi,li2024semgrasp}, without capturing the full spectrum of semantic dimensions relevant to dexterous grasping.

%%%%%%%%%%%%%%%%%%%%%%%%%%%%%%%%%%%%%%%%%%
\section{Methods}
\label{sec:Method}
%%%%%%%%%%%%%%%%%%%%%%%%%%%%%%%%%%%%%%%%%%

\subsection{Problem Statement and Method Overview}
Existing dexterous grasp pose generation methods remain limited in their ability to model semantic information, particularly in scenarios requiring fine-grained control. This limitation restricts further improvements in the precision and adaptability of dexterous grasp strategies. Moreover, the scarcity of grasp datasets with rich, multi-dimensional semantic annotations continues to hinder progress in this research direction.

To address these challenges, we propose a semantic-sensitive dexterous grasp dataset generation method OmniDexDataGen that incorporates multiple levels of semantic information, from functional affordance, contact configuration, to grasp taxonomy,  as introduced in Sec.~\ref{sec:dataset_generation_grasp_taxonomy_aware}. Building upon this foundation, we develop a semantic reasoning pipeline based on large multimodal models OmniDexReasoner to enhance the understanding of object-grasp relationships and grasp type, as shown in Sec.~\ref{sec:dexterous_grasping_semantic_label_generation}. We further introduce a 3D vision-language model for semantically guided dexterous grasp pose generation, named OmniDexGraspNet, as detailed in Sec.~\ref{sec:3d_vision_language_dexterous_grasp_generation}.

\subsection{Functional Affordance-Aware and Contact-Sensitive Dexterous Grasp Dataset Generation Enriched by Grasp Taxonomic Diversity}
\label{sec:dataset_generation_grasp_taxonomy_aware}
We propose a dexterous grasp data generation method that is sensitive to both functional affordance and grasp taxonomy, as shown in Fig.~\ref{fig:grasp_generation_pipeline}. This method integrates the functional regions of the object, the contact configurations, and the opposition types associated with various grasp taxonomies. By leveraging an optimization-based strategy, our approach generates high-quality grasp samples with enhanced functional diversity, contact semantic diversity, and grasp taxonomy diversity. The proposed grasp generation pipeline consists of following key parts: Grasp-type-aware configuration sampling, functional affordance-aware contact point sampling (AC-Sampler), grasp taxonomy-aware DFC grasp pose sampling (Tax-DFCSampler), and grasp optimization method.

In the grasp-type-aware configuration sampling stage, representative grasp configurations are automatically sampled according to the dexterous hand grasp taxonomy. In the affordance- and grasp-type-guided pose sampling method, initial grasp poses are sampled within object affordance regions based on task semantics and grasp-type constraints, including opposition type and corresponding contact configuration. Finally, optimization method refines the sampled grasp poses using functionality- and taxonomy-aware hand–object interaction loss and validates them through physical evaluation to ensure reliable data generation.

\subsubsection{Grasp Taxonomy-Aware Configuration Sampling}

Originating from grasp taxonomy analysis~\cite{feix2015grasp}, different grasp types are typically characterized by specific opposition types, force directions, involved fingers in hand–object contact, contacting links of each finger, and the spatial distribution of contact regions. 
We define a structured representation of a grasp configuration, which includes: the set of activated fingers $S_f$ participating in the grasp, the specific links $S_l$ of each finger that are involved in contact, the contact regions $S_c$ associated with each link and the force direction $v$ and grasp central point $c$.

The goal of grasp configuration sampling is to generate an appropriate grasp configuration $G = \{S_f, S_l, S_c, v, c\}$ given a specified grasp type $t_g$. 

Formally, the grasp configuration sampling process is defined as:
\begin{equation}
    G = f_{\rm config}(t_g)
\end{equation}
where $G$ denotes the generated grasp configuration, and $f_{\rm config}$ is the grasp configuration sampling method conditioned on the specified grasp type.

\subsubsection{Functional Affordance-Aware and Grasp Type-Aware Differential Force-Closure Grasp Sampling}

Building on our proposed grasp configuration modeling, we further introduce a Differential Force Closure (DFC)-based grasp pose sampling method (Tax-DFCSampler) that is sensitive to both functional affordance and grasp type. This method is designed to generate initial grasp pose candidates, enhancing the semantic diversity and physical plausibility of the resulting grasp set. The sampling pipeline is detailed in Alg.~\ref{alg:dfc_gps}. 

The objective of Tax-DFCSampler is to, given a specific grasp configuration, incorporate object-level functional affordance information to generate a set of physically plausible contact points and estimate their corresponding initial grasp poses. The process can be formally defined as:
\begin{equation}
    \{G_{\rm init}, C_{\rm init}\} = f_{\rm Tax-DFCSampler}(M_{\rm obj}, G)
\end{equation}
where, $G_{\rm init}$ is the initial grasp pose. $C_{\rm init}$ denotes the corresponding set of contact points. $M_{\rm obj}$ is the object mesh annotated with an affordance map. $G$ is the grasp configuration and $f_{\rm Tax-DFCSampler}$ is the grasp sampling function.

To enhance sensitivity to functional affordance, functional affordance contact point sampling method (AC-Sampler) first samples two primary contact points $C$  from the affordance map of the object surface. These candidate contacts are then validated using a DFC estimator $f_{\rm DFC}$~\cite{liu2021synthesizing} to ensure that they meet force-closure constraints. 
\begin{equation}
\begin{gathered}
f_{\rm DFC}=\|G c\|^2 \\
G=\left[\begin{array}{ccc}
I_3 & \cdots & I_3 \\
{\left[\psi_1\right]_{\times}} & \cdots & {\left[\psi_n\right]_{\times}}
\end{array}\right] \\
{\left[\psi_k\right]_{\times}=\left[\begin{array}{ccc}
0 & -\psi_k^{(z)} & \psi_k^{(y)} \\
\psi_k^{(z)} & 0 & -\psi_k^{(x)} \\
-\psi_k^{(y)} & \psi_k^{(x)} & 0
\end{array}\right]}
\end{gathered}
\end{equation}
where, $\Psi=\{\psi_1, \cdots,\psi_n\}$ denotes the set of contact point candidates. term $c \in \mathbb{R}^{n \times 3}$ represents the normals of object surface at the contact points in~$\Psi$, and $n$ indicates the number of contact points. 
Specifically, an oversampling strategy is employed, allowing the DFC estimator to filter out physically infeasible candidates and retain a batch-sized set of valid contact points. The mid-point between $C_m$ is selected as the grasp central point for further initial hand pose alignment, and random permutations are added to increase diversity in the resulting grasp poses.

Next, finger joint poses are initialized based on sampled the hand configuration according and the grasp type. 
By default, the initial pose is defined such that the index finger and thumb are positioned in opposition, enabling a neutral pre-grasp configuration commonly used for precision or pad opposition grasps. However, for grasp types that involve side contact regions, the initial hand open pose is specifically adapted to place the thumb in a side opposition configuration. Then, to generate contact points for all activated fingers, all fingertip contact points of all activated fingers are projected along the contact force direction onto the object surface to obtain a full set of candidate contact points. To prevent overfitting to a single structure, the non-activated fingers are randomly perturbed. Following this, we perform parallel hand-object collision checking using the kinematic skeleton of the dexterous hand. Specifically, the generated grasp candidates are split into multiple chunks for efficient parallel evaluation. 
For each pose, the number of collisions between hand links and the object is computed. Only those candidates with a collision count below a predefined threshold are retained as valid initial grasp poses $G_{\rm init}$.

\begin{algorithm}[t]
\caption{Functionality- and Grasp-Type-Aware Grasp Pose Sampling (AC-Sampler + Tax-DFCSampler)}
\label{alg:dfc_gps}
\begin{algorithmic}[1]
\REQUIRE $M_{\text{obj}}$: Object mesh with functional affordance map \\
         $G$: Grasp configuration from taxonomy \\
         $f_{\text{DFC}}$: Differential Force Closure estimator \\
         $N$: Number of required grasp candidates \\
         $\tau$: Collision threshold
\ENSURE $G_{\text{init}}$: Set of valid initial grasp poses

\STATE $C_{\text{valid}} \leftarrow \emptyset$
\WHILE{$|C_{\text{valid}}| < N$}
    \STATE $C_m \leftarrow$ SamplePrimaryContacts($M_{\text{obj}}$, $G$)
    \IF{$f_{\text{DFC}}(C_m)$ is valid}
        \STATE $C_{\text{valid}} \leftarrow C_{\text{valid}} \cup \{C_m\}$
    \ENDIF
\ENDWHILE

\STATE $G_{\text{init}} \leftarrow \emptyset$
\FOR{each $C$ in $C_{\text{valid}}$}
    \STATE $C_m' \leftarrow$ RandomSwap($C_m$)
    \STATE $d \leftarrow$ GetForceDirection($G$)
    \STATE $F_{\text{act}} \leftarrow$ GetActivatedFingers($G$, $d$)
    \STATE $p_{\text{center}} \leftarrow$ EstimateGraspCenter($C_m'$, $d$)
    \STATE $G_{\text{pose}} \leftarrow$ ConstructHandPose($p_{\text{center}}$, $d$, $G$, addNoise= True)
    \STATE RandomizeIdleFingers($G_{\text{pose}}$)
    \STATE $n_{\text{col}} \leftarrow$ CheckCollision($G_{\text{pose}}$, $M_{\text{obj}}$)
    \IF{$n_{\text{col}} < \tau$}
        \STATE $G_{\text{init}} \leftarrow G_{\text{init}} \cup \{(G_{\text{pose}}, C_m')\}$
    \ENDIF
\ENDFOR
\RETURN $G_{\text{init}}$
\end{algorithmic}
\end{algorithm}

\subsubsection{Grasp Optimization and Simulation Validation}

After constructing the grasp configuration that incorporates functional affordance, grasp taxonomy, and contact sensitivity, and obtaining the initial grasp pose through DFC-based sampling, we further refine the candidate poses via an optimization procedure. During optimization, only the activated fingers and wrist pose are updated, while the non-activated fingers remain fixed to their original configuration.
The overall objective function comprises the following three key components: Semantic hand-object interaction loss, Differential Force Closure (DFC) estimation loss and Penetration and joint-limit regularization. 

The construction of the loss function proceeds as follows. 
For each activated finger, we randomly sample contact points from its designated contact region as specified by the grasp configuration. 
Corresponding target contact points are sampled from the object surface. 
The contact alignment is then optimized by computing the Signed Distance Function (SDF) between the hand mesh and object surface, minimizing interpenetration while encouraging semantically valid contact. 
The DFC estimator is employed as an additional optimization objective to promote high-quality, force-closure grasps. 
During the entire optimization process, joint angle updates are constrained by anatomical joint limits, ensuring biomechanically feasible and physically plausible hand postures.

After optimization, each refined grasp candidate is validated within a physics simulation environment. Specifically, we adopt a joint impedance control strategy to enable compliant yet stable dexterous grasping, simulating realistic contact forces and evaluating the stability of the grasp during execution. 
For each joint $i$, the control torque $\tau_{i}$ is computed as:
  \begin{equation}
\tau_i=k_i\left(q_i^{\text {target}}-q_i\right)-d_i \dot{q}_i+g_i
\end{equation}
where, $\tau_{i}$ denotes the torque applied to joint $i$, $q_i$ is the current joint position, $q_i^{\text {target}}$ represents target joint position, $\dot{q}_i$ denotes joint velocity, $k_i$, $d_i$ and $g_i$ are the stiffness, damping coefficients, and external compensation terms, respectively,
Only those grasp poses that successfully complete the simulated grasp—without slippage or failure—are retained in the final dataset.

\begin{figure*}[htbp]
  \centering
  \includegraphics[width=1.0\linewidth]{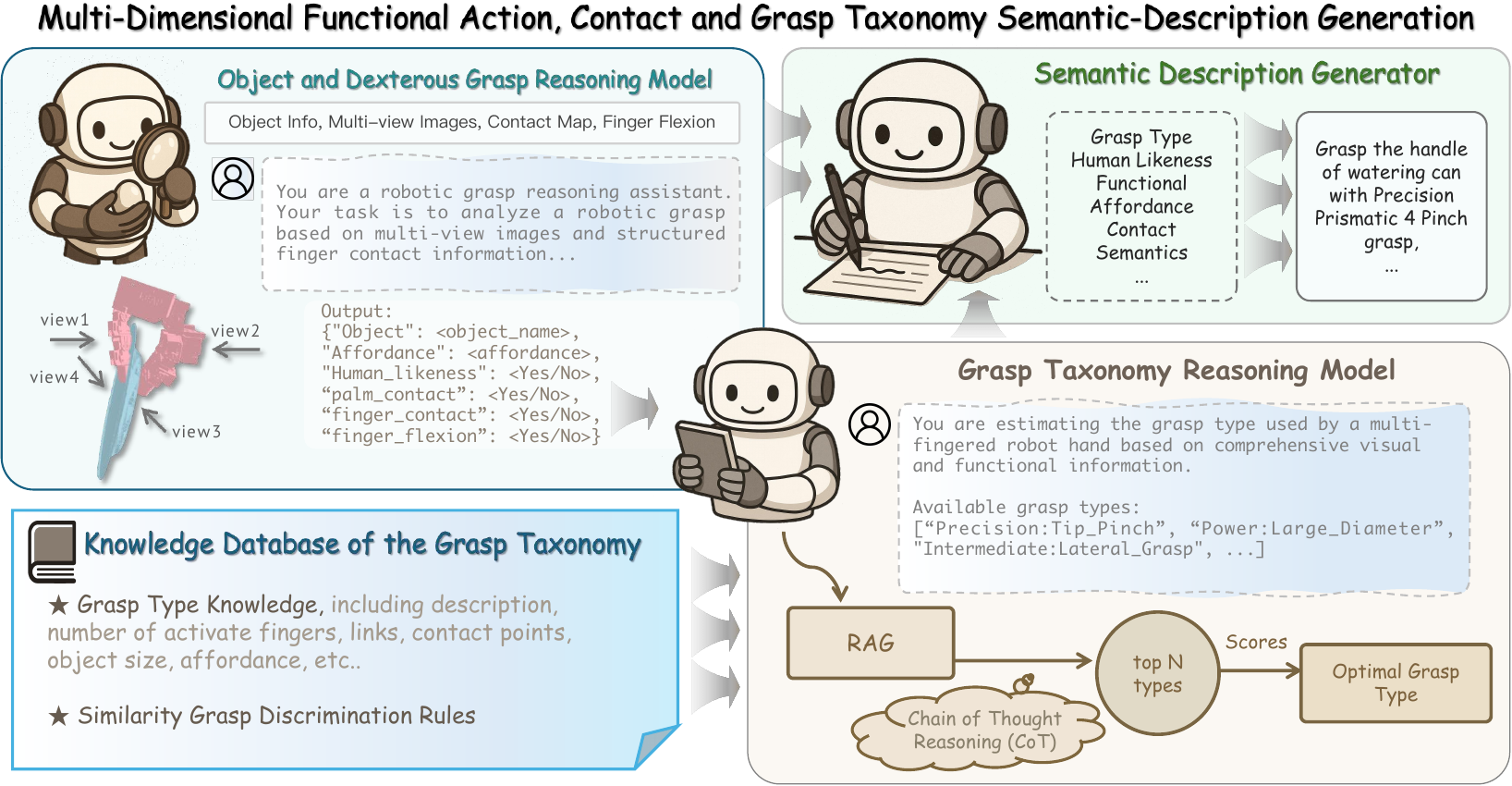}
  \caption{% 
  OmniDexReasoner: 
  LMM-Based dexterous grasping multi-dimensional semantic reasoning, including functional affordance, contact semantic information, finger configuration, grasp taxonomy and human-consistency.}
  \label{fig:semantic_understanding_lmm}
\end{figure*}

\subsection{LMM-Grounded Dexterous Grasping Multi-Dimensional Semantics Understanding Method}
\label{sec:dexterous_grasping_semantic_label_generation}

To enrich the generated grasp dataset with multi-dimensional semantic annotations, we propose OmniDexReasoner, a multimodal large model (LMM)-based semantic reasoning framework for dexterous grasping, which is sensitive to functional affordance, contact semantics, and grasp taxonomy, as shown in Fig.~\ref{fig:semantic_understanding_lmm}. By integrating multi-agent collaboration, Chain-of-Thought (CoT) reasoning, and Retrieval-Augmented Generation (RAG) strategies, the proposed framework addresses key gaps in current research on dexterous hand semantic understanding. 
Our semantic reasoning pipeline consists of three key modules: hand–object interaction understanding model, grasp taxonomy reasoning model and dexterous grasp multi-dimensional semantic generatior.  This framework enhances the expressiveness of semantic annotations in dexterous grasp datasets and provides high-quality semantic priors to support downstream multimodal models.

\subsubsection{Hand–Object Interaction Understanding Model}

To understand the interactive information between a robotic hand and an object, we propose a model for hand-object interaction understanding. This model leverages multimodal information, including object attributes, robotic hand configurations, hand-object contact data, and grasping scene images, to automatically predict the semantic relationships between the hand and the object. 
This process can be modeled in a probabilistic framework as:
\begin{equation}
    P(A|I)=P(A|H,O,C,V)
\end{equation}

where $A$ represents the semantic outputs of the model, and $I$ denotes the multimodal input. $H$ denotes information about the robotic hand, such as the types of available fingers, available finger links, joint configurations, and maximum graspable size. 
$O$ represents object-related information, including the object's name, affordance annotations. $C$ captures the contact information between the hand and the object, including the contact semantic map $M_c$ and corresponding grasp affordance estimation result $a^{*}$. 
$V$ refers to multi-view images of the grasping scene. Output $A$ contains of final functional affordance estimation classification result $a_{\rm sem}$, textual object descriptions, contact fingers and links, finger flexion configurations, palm contact status, and the contact semantic map.

To model the contact information, we compute the spatial distance between various links and fingers of the hand and surface points on the object, similar to~\cite{zhang2024contactdexnet}. This allows us to construct a Contact Semantic Map $M_c$, assigns semantic contact labels to the object surface points. The map is defined as:
\begin{equation}
M_c\left(p_i\right)=\left(l_j, f_k\right), \quad \text { if } \quad d\left(p_i, L_{j k}\right)<\delta
\end{equation}
where, $p_i \in \mathcal{P}$ denotes a surface point sampled from the object. $l_j$ denotes the 
j-th link. $f_k$ denotes the k-th finger. 
$L_{jk}$ is the 3D model of the j-th link on the k-th finger. $d(p_i, L_{jk})$ is the minimum Euclidean distance between $p_i$	
  and $L_{jk}$. $\delta$ is a contact threshold. Each contact point on the object surface is thus labeled with the corresponding hand link and finger involved in the contact.

Within the agent, the Contact Semantic Map is further analyzed to identify the involved fingers and links, and whether the palm is in contact with the object. Palm contact is a key indicator of grasp type, such as power grasp or precision grasp. However, the degree of force exerted by the palm requires further estimation using visual cues from the grasping scene.

To infer object affordances from hand-object interactions, we employ a voting-based functional grasp affordance classification method. This method integrates the Contact Semantic Map $M_c$ with an affordance label map $M_a$ defined over the object surface. For each contact point $p_i$, we count the associated affordance label and determine the most frequent one via voting:
 \begin{equation}
a^*=\arg \max _{a \in \mathcal{A}} \sum_{p_i \in \mathcal{P}} \mathbb{I}\left[M_c\left(p_i\right) \neq \emptyset\right] \cdot \mathbb{I}\left[M_a\left(p_i\right)=a\right]
\end{equation}
where, $a^*$ is the predicted functional affordance. $\mathcal{A}$ is the set of predefined affordance classes (e.g., HandleGrasp, WrapGrasp, Press, Pour, Cut, Stab, Pull, Push, Open, Twist, Hammer, Pry, Support, Lift, Lever, None). $\mathbb{I}[\cdot]$ is the indicator function. $M_c\left(p_i\right) \neq \emptyset$ implies point $p_i$ is in contact with the hand. $M_a(p_i)$ is the affordance label for point. This process selects the dominant affordance type based on the most frequent contact-affordance label co-occurrence.

In addition, we incorporate multi-view images of the grasping scene as part of the input, denoted as the set $\{V\}$. These images contain rich visual information about the physical contact between different parts of the hand and the object, as well as the affordance-relevant features of the interaction. They also reveal the pose of each finger on the robotic hand. The degree of finger flexion jointly reflects the object’s size, the contact force distribution, and the potential grasp strategy. In general, larger objects tend to require power grasps, while smaller or more intricate objects are more likely to involve precision grasps. 

Based on all the aforementioned inputs, the agent is tasked with predicting the semantic attributes of the hand-object interaction, including contact affordances and finger joint configurations, evaluating human-consistency, and then validating the contact understanding derived from conventional methods. More implementation details are introduced in the supplementary materials.

\subsubsection{Grasp Taxonomy Understanding with Multimodal Reasoning}
The grasp type is influenced by a combination of complex factors, including hand-object contact information, force direction, finger joint configurations, and object affordances.  
However, current Large Multimodal Models (LMMs) still suffer from significant hallucination when interpreting physical interactions in the real world, making them unreliable for accurate grasp type prediction. 
To address this challenge, we propose a novel grasp taxonomy semantic understanding model aimed at reasoning about grasp types $T$ from multimodal input signals $I=\{H,O,C,A,V\}$, 
where, $H$, $O$, $C$, $A$, and $V$ represent hand configuration, object properties, hand-object contact information, affordance cues, and multi-view scene images, respectively. 

We formulate grasp taxonomy classification as a two-level hierarchical task from coarse-level classification to fine-level classification. 
Coarse-level classification predicts whether the grasp is Power, Precision, or Intermediate. Fine-level classification: Further classify into specific taxonomy subtypes (e.g., pinch, tripod, hook, etc.).

Grasp taxonomy classification is challenging because of high retrieval and reasoning complexities. The grasp taxonomy space is large and fine-grained, where many grasp types differ only subtly in visual or semantic attributes. This complexity increases the classification difficulty.  Moreover, in different task scenarios, the same object may afford different grasp types depending on human preferences or task-specific goals. This context-dependent variability makes it difficult to classify grasps using physical information alone.
Therefore, it is essential to incorporate both affordance information and semantic reasoning mechanisms to assist grasp type classification. Secondly, task-driven preferences often lead to different grasping strategies even under similar contact conditions, placing high demands on the model's reasoning capabilities. Meanwhile, LMMs frequently generate hallucinated or inconsistent predictions when interpreting physical scenes. To enhance the robustness and reliability of grasp reasoning, external knowledge, semantic context, and multimodal cues must be integrated into the inference pipeline.

To address the challenges mentioned, we incorporate Retrieval-Augmented Generation (RAG) and Chain-of-Thought (CoT) into our grasp taxonomy reasoning framework to enhance the performance of grasp taxonomy reasoning.

The Retrieval-augmented mechanism is introduced to complement the LMM with an external structured knowledge database about the grasp taxonomy. The database is desinged to retrieve relevant semantic and functional information prior to grasp classification. Specifically, the domain-specific knowledge bases consist of a grasp type database and rules about similar grasp discrimination. The grasp type database includes detailed descriptions of all defined grasp types and corresponding semantic knowledge about fingers, links, contacts, and affordance configurations, as well as the use-case examples. This improves the model's ability to ground predictions in contextually relevant knowledge. 
Formally, the model becomes:
\begin{equation}
    T=f_{\rm GTR}(I,{\rm Retrieve}(q))
\end{equation}

where, $f_{\rm GTR}$ is the multimodal model that integrates both inputs and retrievals for final Grasp Taxonomy Reasoning prediction. $q$ is the query constructed from $I$. 
$Retrieve(q)$ denotes the retrieved context from the knowledge database. 

Chain-of-Thought (CoT) prompting is also incorporated to perform intermediate reasoning steps before arriving at a final grasp type prediction. This enhances interpretability and logical consistency in inference. The reasoning chain is structured as: grasp scene, affordance inference, contact type identification, grasp taxonomy prediction.

\subsubsection{Dexterous Grasp Multi-Dimensional Semantic Description Generation Model}

To enhance the semantic interpretability of dexterous grasp behaviors and integrate multi-modal and multi-level semantic information, we propose a Dexterous Grasp Descriptor Generator. This module is designed to produce natural language descriptions that characterize grasp actions by jointly reasoning over key semantic components embedded in the grasping process.

The generator takes as input a combination of semantic signals derived from the grasping interaction, including the functional affordance of the object, the contact semantics between the hand and the object, and the associated grasp taxonomy label. By modeling the interdependence between these factors, the generator is capable of composing diverse and contextually appropriate textual descriptions that reflect the nature of the grasp. For instance, “Two fingertips pinch the narrow handle, performing a precision grasp adapted for fine manipulation tasks.”

This descriptor generation module enhances the interpretability of the grasping process by establishing a semantic bridge between physical interaction and linguistic representation. As well, it enables the language-guided grasping pose generation network training, where a natural language interface is essential for aligning robotic actions with human intent.

\begin{figure*}[htbp]
  \centering
  \includegraphics[width=1.0\linewidth]{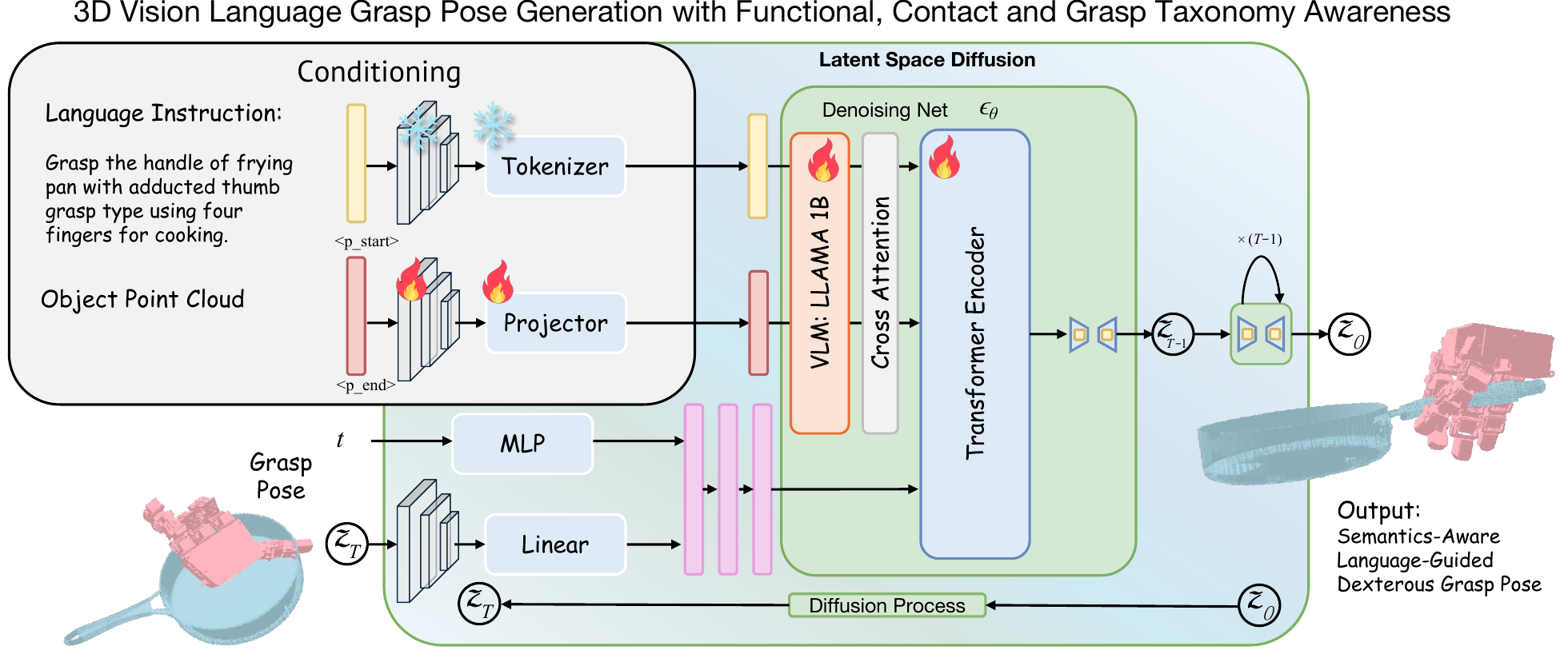}
  \caption{%
  OmniDexGraspNet: Vision Language Dexterous Hand Grasping Pose Generation Model. 
  The textual task description and the partial point cloud of the object are encoded into their respective latent features using a text encoder and a point cloud encoder. Simultaneously, the initial grasp pose is mapped into the latent space via a multilayer perceptron (MLP). These multimodal latent features are then fused and processed by a diffusion transformer encoder, which is trained to predict and remove noise in the grasp pose representation. This denoising process refines the grasp configuration to produce a physically plausible and task-relevant hand pose following language instruction. The input text instructions include key semantic elements such as the grasp affordance location, the intended grasp type, contact fonfiguration, and the high-level task objective.
  }
  \label{fig:network_structure}
\end{figure*}

\subsection{Vision Language Dexterous Hand Grasping Pose Action Generation Model}
\label{sec:3d_vision_language_dexterous_grasp_generation}

To enable effective perception and response to multi-dimensional semantic cues in dexterous grasping tasks, we propose a language-conditioned Vision-Language Grasping Generation (VLG) model that generates semantically aligned dexterous grasp poses $\mathbf{g} \in \mathbb{R}^d$ based on partial point cloud observations $P \in \mathbb{R}^{N \times 3}$ of the target object and language instruction $\mathcal{L}$, as shown in Fig.~\ref{fig:network_structure}. 
The model is capable of modeling and leveraging three key semantic dimensions: functional affordance, contact semantics, and grasp taxonomy, allowing it to produce grasp actions that are highly consistent with the task-specific semantic intent. 

The proposed network comprises the key components: 
Multimodal Conditioning Embedding 
and Grasp Pose Diffusion Generator. 
The input language instruction and point cloud are embedded into a shared semantic space via a Vision-Language Model (VLM):
\begin{equation}
\mathbf{c}=f_{\mathrm{VLM}}(\mathcal{L}, P)
\end{equation}
where $f_{\rm VLM}$ denotes a pretrained transformer-based model (e.g., LLaMA-1B) with cross-attention between the language and visual tokens. The output $c \in \mathbb{R}^{d_c}$ serves as a semantic condition for pose generation.

For aligning the vision-language model with point cloud encoder, the PointNet-based encoder, projector are trained following the training strategy from PointLLM~\cite{xu2025pointllm,xu2024pointllm}. During first training stage, the projector is trained using the multimodal alignment dataset with point clouds, images and natural language annotations~\cite{xu2025pointllm,xu2024pointllm}. The point cloud encoder extracts local geometric features $f_{\rm pcd}$ from partial object point clouds. The extracted features are passed through a projector that maps them into a shared multimodal embedding space to enable semantic alignment with language features from language tokenizer and encoder. 

The multimodal feature fusion module captures complex cross-modal relationships between visual geometry and linguistic semantics, enabling the model to understand what to grasp, how to grasp. 

Grasp pose diffusion transformer generator is introduced to generate semantically aligned grasp poses from semantic condition $c$. 
The grasp pose is represented as a latent variable $z_{\theta}$, generated by gradually denoising a Gaussian-initialized latent $\mathbf{z}_T \sim \mathcal{N}(0, I)$ through a conditional denoising network:

\begin{equation}
\mathbf{z}_0=\mathbf{z}_T-\sum_{t=1}^T \epsilon_\theta\left(\mathbf{z}_t, \mathbf{c}, t\right)
\end{equation}
where $\epsilon_\theta$ is the noise prediction network , conditioned on timestep $t$ and semantic features $\mathbf{c}$. 

The training pipeline consists of two main stages. After training projector for point cloud understanding, pose generator is trained with generated dexterous grasp dataset annotated with rich semantic labels based on proposed OmniDexDataGen and OmniDexReasoner. The OmniDexGraspNet is trained to generate high-quality grasp poses that are sensitive to multiple semantic conditions and adaptable to different language instructions.

%%%%%%%%%%%%%%%%%%%%%%%%%%%%%%%%%%%%%%%%%%
\section{Experiments}
\label{sec:Experiment}
%%%%%%%%%%%%%%%%%%%%%%%%%%%%%%%%%%%%%%%%%%
\subsection{Experimental Setup}

\begin{figure}
    \centering
    \includegraphics[width=1.0\linewidth]{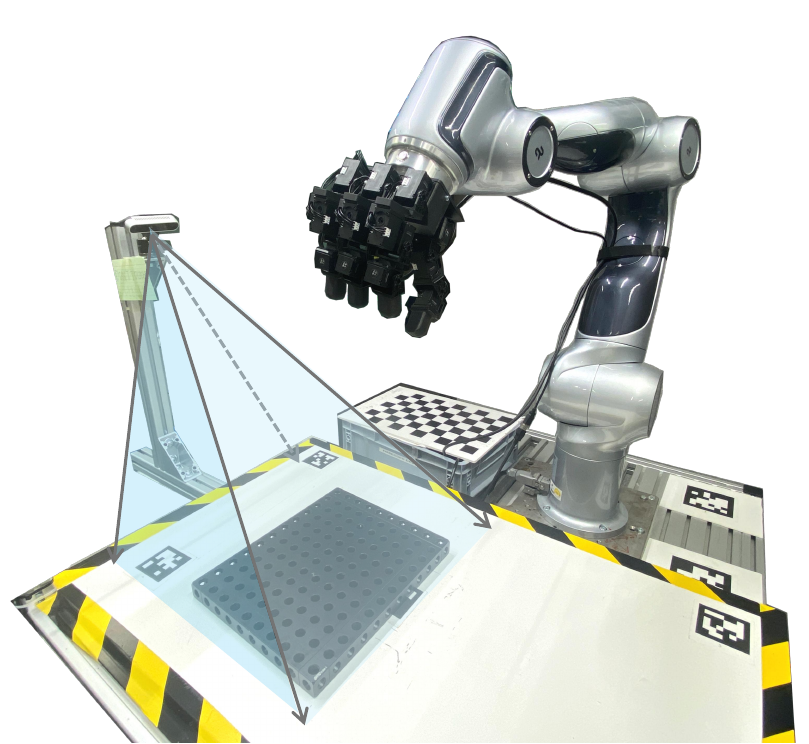}
    \caption{Real world experimental setup with 7-axis Diana robotic arm, LEAP hand and RealSense D415 3D camera.}
    \label{fig:exp_setup_real_world}
\end{figure}

Real world experiment setup is shown in Fig.~\ref{fig:exp_setup_real_world}. We validate our proposed grasping approach on a real robotic platform composed of a Diana robotic arm and a LEAP Hand dexterous manipulator. During execution, the system first generates semantically-informed grasp poses using our method. Grasp trajectory planning is then performed via a constraint-based optimization algorithm, implemented using the PyRoki motion planning library~\cite{kim2025pyroki}. Finally, the robot executes the planned grasp through impedance control, ensuring compliant and stable interaction during the grasping process.

To conduct controlled simulation experiments on grasp dataset generation, semantic reasoning, and action generation, we utilize both the LEAP hand and the DLR-HIT II hand in Isaac Gym simulation environments. 
These settings allow us to evaluate the generalizability and performance of the proposed dataset generation method OmniDexDataGen and semantic reasoning method OmniDexReasoner across different hand hardware. 
The simulation experiments include the following components: Grasp Dataset Generation Evaluation, Multi-Dimensional Semantic Understanding Module Evaluation, Semantic-Conditioned Grasp Action Generation Evaluation. OmniDexGraspNet is evaluated in simulation and real world using LEAP hand.

\subsection{Experiments of Grasp Dataset Generation Method}

\subsubsection{Qualitative Experiment}
The qualitative evaluation of OmniDexDataGen includes comparisons against state-of-the-art grasp dataset generation methods~\cite{wang2022dexgraspnet}, as well as an ablation study across several variant configurations of our approach. The evaluated baselines and ablations are summarized as follows:
\begin{itemize}
    \item \textsc{Baseline 1}: Optimization-based dexterous grasp generaion (OGG),  DexGraspNet~\cite{wang2022dexgraspnet}, which employs global contact point sampling without any semantic awareness and guidance.
    \item \textsc{Baseline 2}: OGG + Grasp taxonomy-aware grasp sampling (Tax-DFCSampler)
    \item \textsc{Baseline 3}: OGG + Affordance-aware contact point sampling (AC-Sampler).
    \item Our OmniDexGen with both AC-Sampler and Tax-DFCSampler.
\end{itemize}

\begin{figure*}[htbp]
  \centering
  \includegraphics[width=1.0\linewidth]{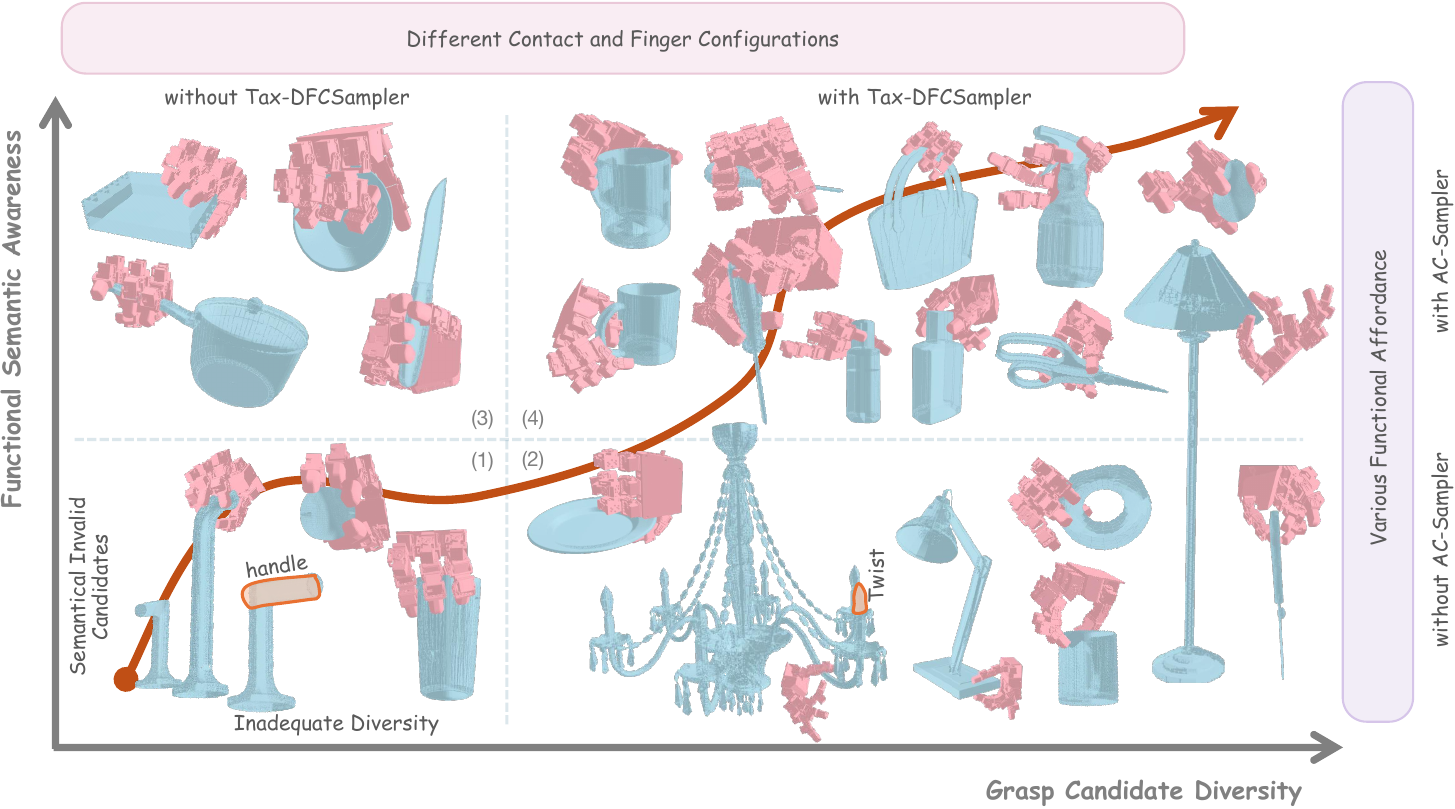}
  \caption{Grasp generation results using state-of-the-art and different variants of our method. (1) DexGraspNet~\cite{wang2022dexgraspnet}, (2) OGG + Tax-DFCSampler, (3) OGG + AC-Sampler, (4) Ours: OGG + AC-Sampler + Tax-DFCSampler.}
  \label{fig:result_grasp_dataset_generation_results}
\end{figure*}

\begin{table*}[tb!]
  \caption{Comparison experimental results of proposed functional affordance-aware and grasp type-aware dexterous grasp dataset generation method with state-of-the-art and different variants.}
  \label{tab:results_dataset_generation}
  \centering
  \setlength{\tabcolsep}{1.15mm}
  \resizebox{1.0\textwidth}{!}{
  \begin{tabular}{lccccccc}
    \toprule
    \multicolumn{1}{l}{\multirow{2}{*}{\textbf{Model}}} & 
    \multicolumn{4}{c}{\textbf{Grasp Diversity}} 
    & \textbf{Contact Surface Coverage} 
    & \textbf{Affordance Diversity}
    \\
    \cmidrule(r){2-5}  \cmidrule(r){6-6} \cmidrule(r){7-7} 
    & KL$^{\downarrow}$ & STD$_{\rm translation}$$^{\uparrow}$ & STD$_{\rm orientation}$$^{\uparrow}$ & STD$_{\rm joint}$$^{\uparrow}$ & Average Hausdorff Distance (CM)$^{\downarrow}$& KL$^{\downarrow}$\\  
\midrule
DexGraspNet~\cite{wang2022dexgraspnet} & $0.259$ & $7.834$ & $0.713$ & $0.921$ & $36.634$ & $0.288$\\
    Ours wo Affordance Contact Sampler & $0.083$ & $9.274$ & $1.057$ & $1.328$ & $35.498$ & $0.109$\\
    Ours wo grasp taxonomy-aware DFC sampler & $0.209$ & $9.987$ & $1.182$ & $1.053$ & $26.173$ & $0.214$ \\
    \rowcolor{blue!10}
    Ours: OGG + AC-Sampler + Tax-DFCSampler & $\mathbf{0.047}$ & $\mathbf{11.709}$ & $\mathbf{1.413}$ & $\mathbf{1.623}$ & $\mathbf{25.518}$ & $\mathbf{0.055}$ \\
    \midrule
  \end{tabular}
  }
\end{table*}

\begin{figure}[htbp]
  \centering
  \includegraphics[width=0.9\linewidth]{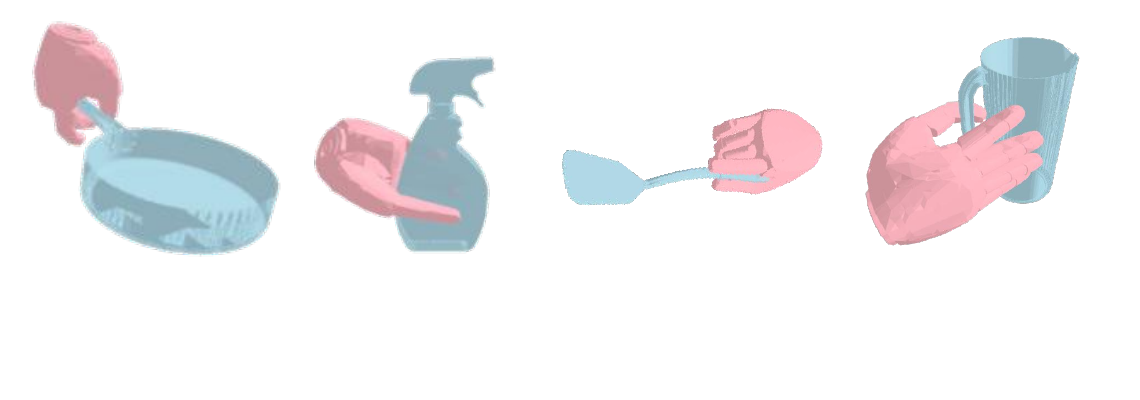}
  \caption{Grasp generation results with DLR-HIT II hand.}
\label{fig:result_grasp_dataset_generation_results_hithand}
\end{figure}

The qualitative results are summarized in Fig.~\ref{fig:result_grasp_dataset_generation_results} with leap hand and Fig.~\ref{fig:result_grasp_dataset_generation_results_hithand} with DLR-HIT II hand. 
Overall, our method demonstrates superior performance across multiple aspects of grasp generation, including controllability over functional affordance, diversity of contact configurations, variation in finger joint configurations, and grasp taxonomy coverage. 

In contrast, \textsc{Baseline 1} performs poorly across all these dimensions, lacking both semantic awareness and structural diversity. Lack of link- and contact region-aware contact sampling also leads to spatial discontinuity of sampled contact points and unstable pose optimization, especially when applied to large-scale objects, as shown in Fig.~\ref{fig:result_grasp_dataset_generation_results} (1). 

\textsc{Baseline 2}, while incorporating improvements over the original version, does not explicitly model functional affordance sampling. As a result, it often generates grasps that are semantically misaligned—for instance, grasping the main body of a refrigerator rather than its handle. Our approach with AC-Sampler supports adaptive grasp generation across objects of varying sizes. Representative samples generated by our approach are illustrated in the Fig.~\ref{fig:result_grasp_dataset_generation_results} (2). 

\textsc{Baseline 3} removes the taxonomy-aware DFC sampler, as shown in Fig.~\ref{fig:result_grasp_dataset_generation_results} (3), leading to limited variation in contact regions and finger configurations. Consequently, the generated grasps are heavily biased towards multi-finger pinch types, lacking diversity across the grasp taxonomy spectrum.

These qualitative observations highlight the importance of jointly modeling functional intent, contact semantics, and grasp type priors in achieving semantically grounded and physically diverse dexterous grasp generation, as shown in Fig.~\ref{fig:result_grasp_dataset_generation_results} (4).

\subsubsection{Evaluation Metrics for Quantitative Experiments} To comprehensively evaluate the quality and semantic diversity of the generated grasp candidates, we adopt a set of metrics that quantify grasp pose variability, contact surface coverage, and functional affordance diversity. Specifically, we use Standard deviations (STDs) of hand position, orientation and joint angles and Kullback–Leibler (KL) divergence to assess pose-level diversity, average Hausdorff distance to measure diversity in contact semantics, and an affordance-level KL divergence to evaluate semantic diversity in grasp functionality.

Using KL divergence calculation, we estimate the diversity of generated categorical data—such as grasp types or functional affordances—by comparing the model's output distribution to a uniform reference distribution, which represents the ideal diversity. Given a total of $N$ categories, the ideal distribution is defined as:
\begin{equation}
P_{\text {ideal }}=\left[\frac{1}{N}, \frac{1}{N}, \ldots, \frac{1}{N}\right]
\end{equation}

Let $P_{\mathrm{gen}}=\left[p_1, p_2, \ldots, p_N\right]$ be the empirical distribution over categories produced by the model. The Kullback–Leibler (KL) divergence from the uniform distribution is computed as:
\begin{equation}
\mathrm{KL}\left(P_{\text {gen }} \| P_{\text {ideal }}\right)=\sum_{i=1}^N p_i \log \left(\frac{p_i}{1 / N}\right)
\end{equation}
A lower KL divergence indicates that the generated distribution is closer to uniform, implying higher diversity across categories. This evaluation approach is applicable to both grasp type diversity and functional affordance distribution diversity.

Specifically, average hausdorff distance (AHD) is calculated using contact semantic maps of generated grasp candidates to quantify contact surface coverage. 
Given a set of $N$ generated grasps in each object, we denote the contact semantic maps as point sets $\{C_1, C_2,\dots, C_{N}\}$. The pairwise average Hausdorff distance between each unique pair $C_i, C_j$ is computed as:
 \begin{equation}
\operatorname{AHD}\left(C_i, C_j\right)=\frac{1}{\left|C_i\right|} \sum_{x \in C_i} \min _{y \in C_j}\|x-y\|+\frac{1}{\left|C_j\right|} \sum_{y \in C_j} \min _{x \in C_i}\|y-x\|
\end{equation}
The final contact surface coverage score is defined as the average AHD across all unique grasp pairs:
\begin{equation}
\text { Score }_{\mathrm{AHD}}=\frac{2}{N(N-1)} \sum_{i<j} \operatorname{AHD}\left(C_i, C_j\right)
\end{equation}

A higher AHD indicates a wider spatial distribution of contact points across the object surface, reflecting the model’s ability to explore a broader range of feasible grasping regions, rather than its sensitivity to functional semantics.

\subsubsection{Quantitative Experiments and Results}
In the quantitative experiments for grasp generation, we evaluate the generated grasp candidates across above key metrics. 
The quantitative results are summarized in Tab.~\ref{tab:results_dataset_generation}.

Quantitative results in Tab.~\ref{tab:results_dataset_generation} demonstrate that the proposed combination of the AC-Sampler and Tax-DFCSampler improves the diversity of the generated grasp dataset. Specifically, the full model achieves the best performance across all grasp diversity metrics, including translation, rotation, and articulation, as indicated by the highest standard deviations and the lowest KL divergence.  
Moreover, the average Hausdorff distance shows that this combination also leads to better coverage of contact semantic patterns. 
The KL divergence of affordance distributions is further reduced, indicating that the generated samples exhibit richer functional affordance variations compared to all baselines and ablations. 

\subsubsection{Analysis and Discussion}
By introducing functional affordance-driven contact sampling, our method substantially enhances the semantic alignment between generated grasp poses and the intended functional regions of the object. Furthermore, the integration of grasp taxonomy-aware DFC  sampling enables the generation of grasps across a wide range of grasp type, each characterized by distinct finger combinations, link involvement, and contact configurations.

Our method achieves superior performance in terms of grasp type diversity, functional affordance expressiveness, and contact semantic diversity.

\subsubsection{Limitation of OmniDexGen.} Despite the capability to generate diverse grasp types, the visual and structural similarity between many grasp types poses challenges for straightforward classification. Existing off-the-shelf classifiers struggle to distinguish between subtle variations. Then, OmniDexReasoner is proposed to automatically infer the semantic category of generated grasps, enabling better annotation and downstream evaluation.

\begin{figure*}[htbp]
  \centering
  \includegraphics[width=1.0\linewidth]{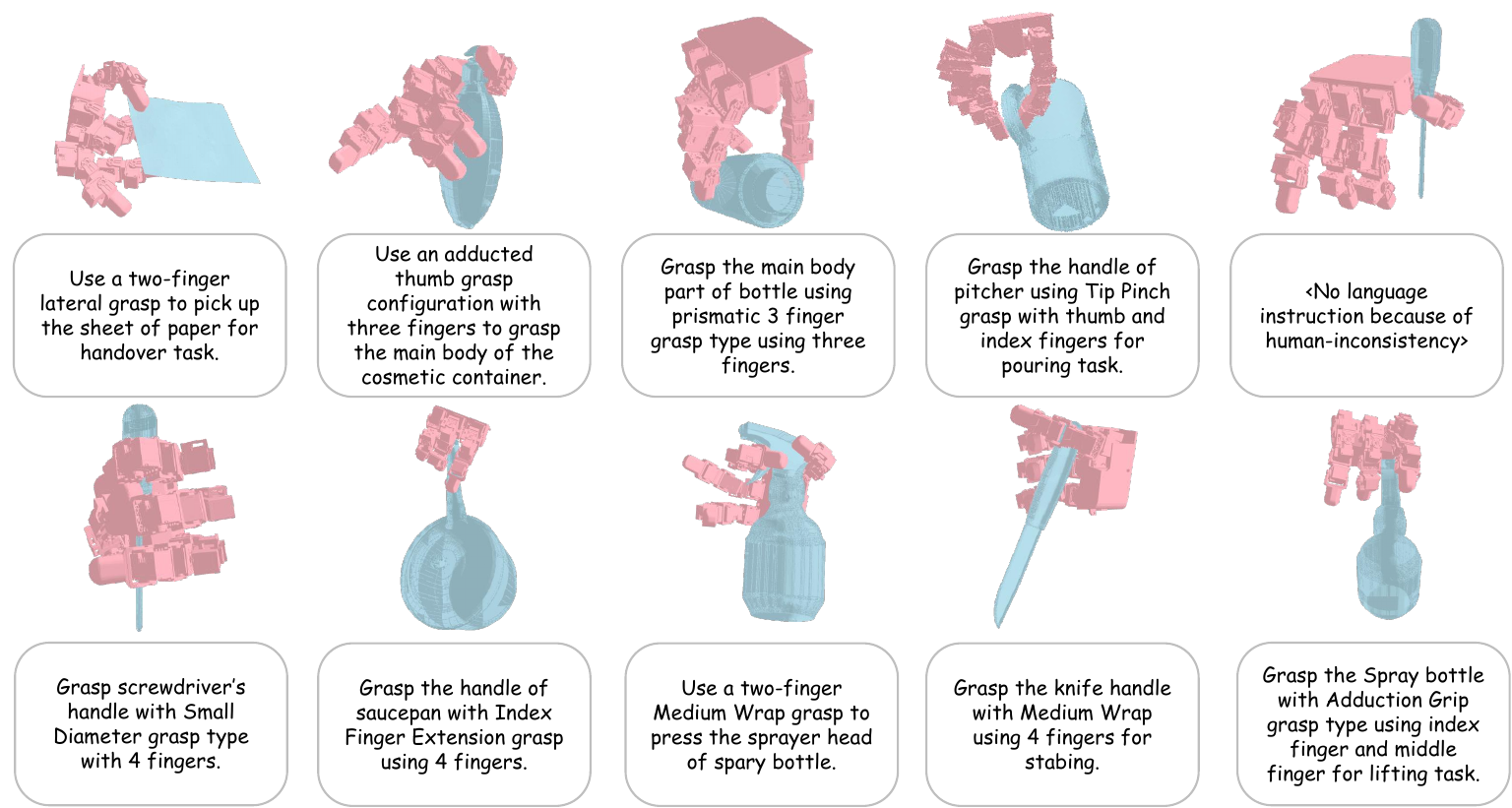}
  \caption{Results of dexterous grasp semantic understanding model OmniDexReasoner.}
\label{fig:results_omnidexreasoner_examples}
\end{figure*}

\begin{table*}[tb!]
  \caption{Quantitative results of OmniDexReasoner compared with state-of-the-art baselines and its ablated variants. $ACC_{\rm coarse}$: classification accuracy of coarse grasp types (power/intermediate/precision). $ACC_{\rm fine}$: classification accuracy of fine grasp types. $ACC_{\rm human}$: classification accuracy of human-like grasp candidates. $ACC_{\rm finger}$: classification accuracy of finger contact status. ${ACC_{\rm func}}$: classification accuracy of  functional affordance.}
  \label{tab:results_different_instruction_types}
  \centering
  \setlength{\tabcolsep}{1.15mm}
  \resizebox{0.8\textwidth}{!}{%
  \begin{tabular}{lcccccc}
    \toprule
    \multicolumn{1}{l}{\multirow{2}{*}{\textbf{Model}}} & 
    \multicolumn{3}{c}{\textbf{Grasp Taxonomy}} &\textbf{Contact Semantic} 
    & \textbf{Functional Affordance} \\
    % \multicolumn{4}{c}{\textbf{Attack}} & 
    % \multicolumn{4}{c}{\textbf{Defense}} \\
    \cmidrule(r){2-4} \cmidrule(r){5-5} \cmidrule(r){6-6} 
    % \cmidrule(r){6-9} \cmidrule(r){10-13} 
    & $ACC_{\rm coarse}$$^{\uparrow}$ & $ACC_{\rm fine}$$^{\uparrow}$ & $ACC_{\rm human}$$^{\uparrow}$ & $ACC_{\rm contact}$$^{\uparrow}$ & $ACC_{\rm func}$$^{\uparrow}$\\
\midrule
    % \rowcolor{light-blue}
    \rowcolor{gray!20}
\multicolumn{6}{l}{\textit{\textbf{Effectiveness of Multi-Agent Collaboration}}} \\ 
    $\pi_{\rm single}$ & $0.52$ & $0.23$ & $0.23$ & $0.36$ & $0.24$\\
    \rowcolor{blue!10}
    $\pi_{\rm multi}$ & $\mathbf{0.89}$ & $\mathbf{0.76}$ & $\mathbf{0.59}$ & $\mathbf{0.68}$ & $\mathbf{0.63}$\\
    \midrule
    \rowcolor{gray!20}\multicolumn{6}{l}{\textit{\textbf{Ablation Study of Key Semantic Modalities.}}} \\ 
    \textsc{$\pi_{\rm multi}$ wo Vision} & $0.84$ & $0.75$ & \cellcolor{yellow!20} $0.34$ & $0.61$ & $0.58$\\
    \textsc{$\pi_{\rm multi}$ wo Finger Info} & $0.88$ & \cellcolor{yellow!20}$0.51$ & $0.58$ & $0.69$ & $0.61$\\
    
    \textsc{$\pi_{\rm multi}$ wo Contact Info} & \cellcolor{yellow!20}$0.47$ & \cellcolor{yellow!20}$0.45$ & $0.59$ & \cellcolor{yellow!20}$0.23$ & $0.66$\\
    \textsc{$\pi_{\rm multi}$ wo Affordance Info} & $0.74$ & $0.70$ & $0.57$ & $0.66$ & \cellcolor{yellow!20}$0.22$\\
    \midrule
    \rowcolor{gray!20}\multicolumn{6}{l}{\textit{\textbf{Effectiveness of Chain-of-Thought and and Retrieval-Augmented Generation.}}} \\ 
    \textsc{$\pi_{\rm multi}$ + CoT} & $0.95$ & $0.80$ & $0.65$ & $0.75$ & $0.63$ &\\
    \textsc{$\pi_{\rm multi}$ + RaG} & $0.90$ & $0.77$ & $0.60$ & $0.71$ & $0.67$ &\\
    \rowcolor{blue!10}
    \textsc{$\pi_{\rm multi}$ + CoT + RaG} (Ours) & $\mathbf{0.97}$ & $\mathbf{0.83}$ & $\mathbf{0.68}$ & $\mathbf{0.77}$ & $\mathbf{0.70}$ &\\
    \midrule
    \rowcolor{gray!20}\multicolumn{6}{l}{\textit{\textbf{Comparison experiments using different LMMs.}}}\\
    \textsc{Qwen7B} & $0.73$ & $0.33$ & $0.49$ & $0.39$ & $0.49$ &\\
    \textsc{Qwen72B} (Ours) & $\mathbf{0.97}$ & $0.83$ & $0.68$ & $0.77$ & $0.70$ &\\
    \textsc{GPT4o-mini} & $0.95$ & $0.85$ & $0.73$ & $0.79$ & $0.74$ &\\
    \rowcolor{blue!10}
    \textsc{GPT4o} & $\mathbf{0.97}$ & $\mathbf{0.92}$ & $\mathbf{0.84}$ & $\mathbf{0.82}$ & $\mathbf{0.78}$ &\\
    \midrule
  \end{tabular}}
\end{table*}

\subsection{Experiment of LMM-Based Dexterous Grasping Semantics Understanding Method}

\subsubsection{Qualitative Experiments}

In the qualitative experiments, we demonstrate the capability of the proposed semantic reasoning module to interpret multiple grasp pose candidates across multiple semantic dimensions, as illustrated in Fig.~\ref{fig:results_omnidexreasoner_examples}. Inference results along three core semantic dimensions: contact semantics, grasp taxonomy, and functional affordance. The visualization results show that our model can generate structurally coherent and semantically consistent grasp descriptions, exhibiting a strong ability to capture the interdependence among different semantic dimensions.

\subsubsection{Quantitative Experiments}

To quantitatively evaluate the reasoning accuracy, we construct a manually annotated benchmark dataset consisting of 100 grasp poses. Each pose is labeled with ground truth annotations with contact configuration, coarse and fine grasp types, and functional affordance. 
This annotated set serves as the Ground Truth (GT) for evaluation. Based on this GT, we compute multidimensional classification accuracies, including coarse and fine grasp taxonomy classification accuracies ($ACC_{\rm coarse}$ and $ACC_{\rm fine}$), human-consistency accuracy $ACC_{\rm human}$, contact semantics classification accuracy $ACC_{\rm contact}$, and functional affordance classification accuracy $ACC_{\rm func}$. 

The experiment results of variants using different agent configurations, key semantic modalities, various LMMs are shown in Tab.~\ref{tab:results_different_instruction_types}. The experiments with different agent configurations and key semantic modalities are executed with Qwen 72B~\cite{bai2025qwen2}. Comparison experiments using different LLMs are finished with optimal configuration using Qwen 7B, Qwen 72B~\cite{bai2025qwen2}, GPT4o-mini and GPT4o~\cite{hurst2024gpt}.

\subsubsection{Ablation Study Results and Analysis}
Results demonstrate that the multi-agent variant substantially outperforms its single-agent counterpart, achieving notable improvements in $ACC_{\rm coarse}$ (0.89), $ACC_{\rm contact}$ (0.68), and $ACC_{\rm func}$ (0.63). This highlights the advantage of collaborative reasoning in multi-agent settings for complex embodied tasks.

Ablation studies based on Qwen72B further reveal the critical role of multi-modal information. Removing visual inputs severely degrades $ACC_{\rm human}$, while the absence of finger contact cues mainly impacts $ACC_{\rm fine}$. Contact semantics are essential for $ACC_{\rm contact}$, and affordance-related inputs are key to functional reasoning, as evidenced by the sharp drop in $ACC_{\rm func}$ when removed. These findings underscore the necessity of integrating diverse semantic modalities for robust performance.

Moreover, incorporating chain-of-thought (CoT) and retrieval-augmented generation (RAG) mechanisms leads to consistent improvements across all metrics. The combination of CoT and RAG yields the best results, suggesting that multi-step reasoning and external knowledge retrieval are complementary in enhancing fine-grained grasp understanding and affordance inference.

Finally, cross-model comparisons show that GPT-4o achieves state-of-the-art performance, outperforming Qwen72B and GPT-4o-mini across all evaluation dimensions. This confirms the scalability and generalization capability of the proposed reasoning framework when paired with advanced large language models.

\begin{figure*}[htbp]
  \centering
  \includegraphics[width=1.0\linewidth]{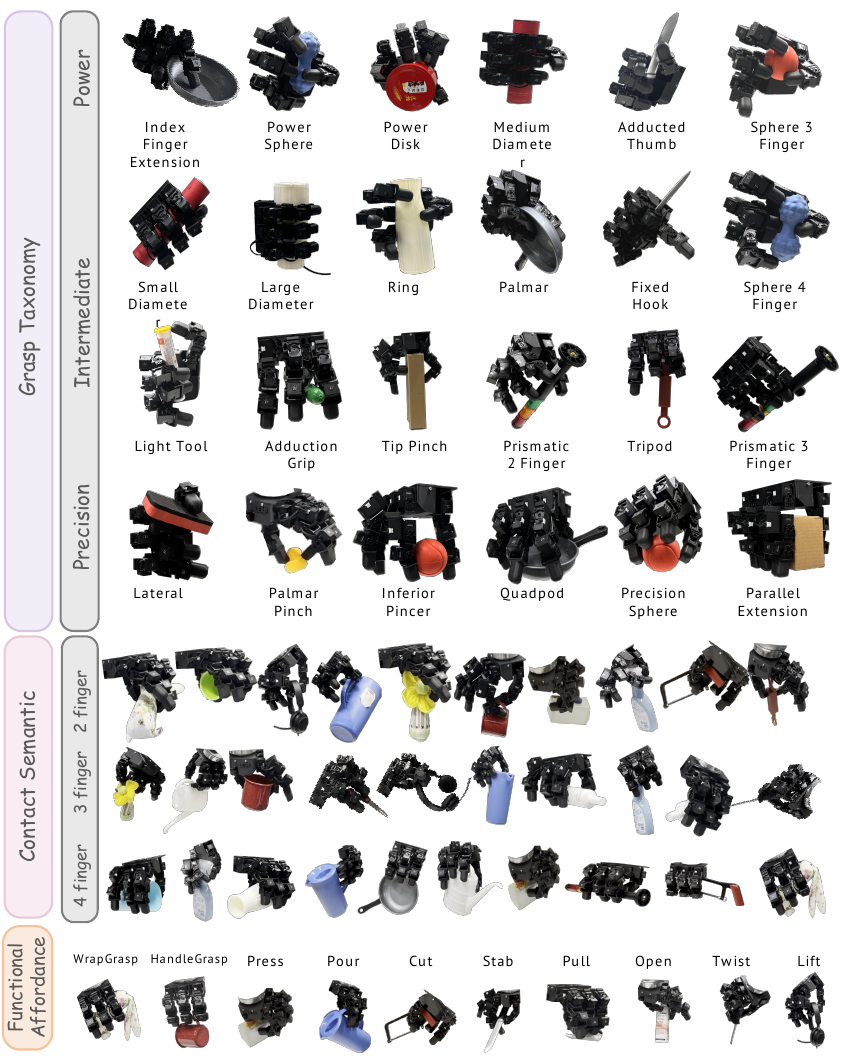}
  \caption{% 
  Grasp taxonomy, contact semantic and functional affordance distribution of dexterous grasp generation.}
  \label{fig:result_grasp_taxonomy_real_world}
\end{figure*}

\subsection{Experiments of 3D Vision-Language Dexterous Grasp Pose Generation Network}

\subsubsection{Simulation Comparison Experiments}
We conduct comparative grasping experiments to further evaluate the effectiveness of our grasp generation framework OmniDexGraspNet. In each trial, the model is provided with a partial point cloud of the target object along with a corresponding language instruction that specifies the intended grasp semantics. The estimated grasp poses are validated in the simulation environment by determining whether the object can be held stably. In addition to grasp success rate, we also record the resulting contact configuration, grasp taxonomy, and functional affordance of each successful grasp to assess the semantic consistency and diversity of the generated behaviors. 
Comparisons are made between our proposed method and existing state-of-the-art baselines with 100 grasping trials. The results are summarized in Tab.~\ref{tab:results_simulation_comparison_grasping_exp}.

\begin{figure*}[htbp]
  \centering
\includegraphics[width=0.9\linewidth]{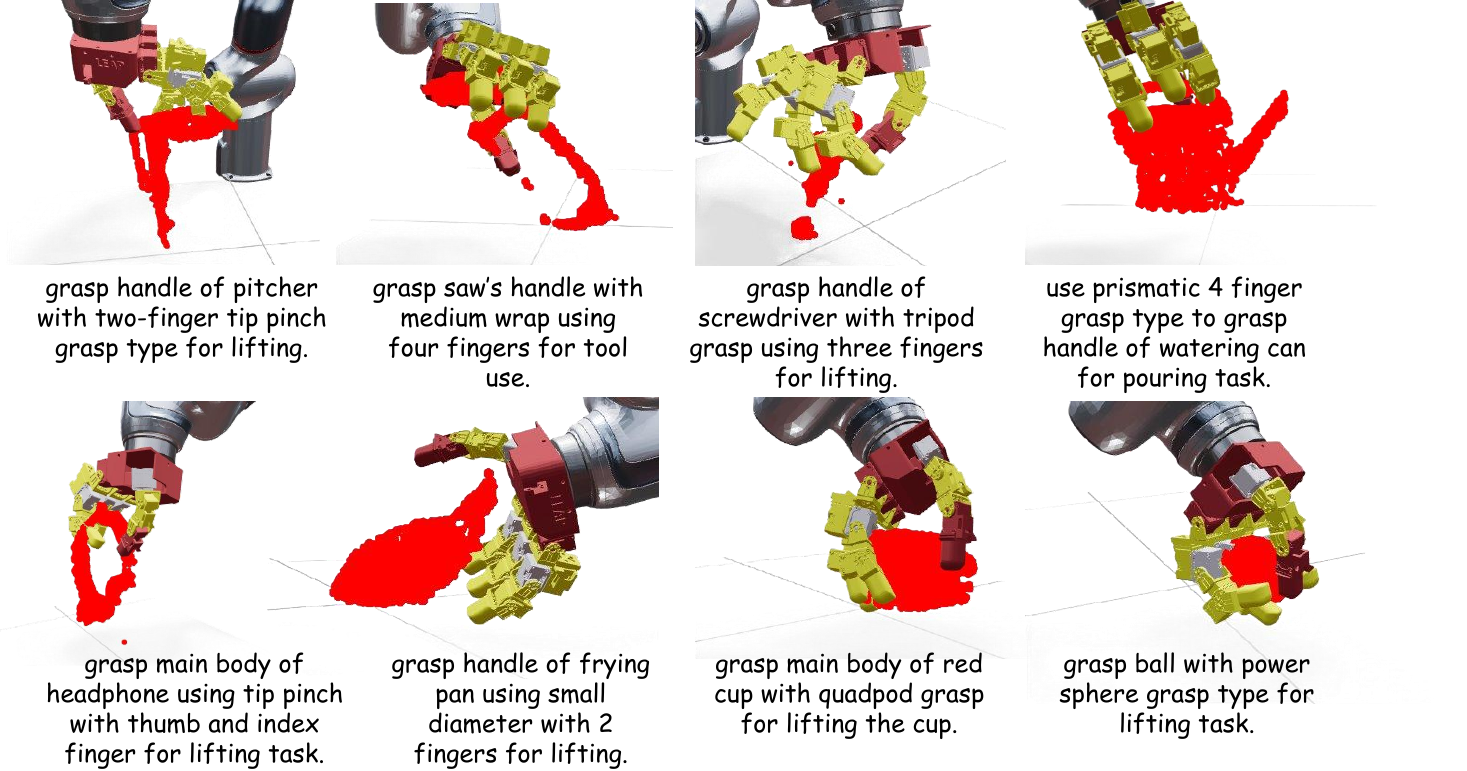}
  \caption{Real-world inference results using proposed OmniDexGraspNet.}\label{fig:inference_result_omnidexgraspnet}
\end{figure*}

\begin{table}[tb!]
  \caption{Semantic information-guided grasping generation experiments results in simulation and real world. GSR: Grasp success rate, $ACC_{\rm FC}$: Finger contact semantic accuracy, $ACC_{\rm GT}$: Grasp taxonomy semantic accuracy, $ACC_{\rm FA}$: Functional grasp affordance semantic accuracy.}
  \label{tab:results_simulation_comparison_grasping_exp}
  \centering
  \setlength{\tabcolsep}{1.15mm}
  \resizebox{0.48\textwidth}{!}{%
  \begin{tabular}{lcccccccc}
    \toprule
    \multicolumn{1}{l}{\multirow{2}{*}{\textbf{Model}}} & 
    \multicolumn{4}{c}{\textbf{Simulation}} 
    & \multicolumn{4}{c}{\textbf{Real-World}} \\
    \cmidrule(r){2-5}  \cmidrule(r){6-9}
    & GSR$^{\uparrow}$ & $ACC_{\rm FC}$$^{\uparrow}$ & $ACC_{\rm GT}$$^{\uparrow}$ & $ACC_{\rm FA}$$^{\uparrow}$ 
    & GSR$^{\uparrow}$ & $ACC_{\rm FC}$$^{\uparrow}$ & $ACC_{\rm GT}$$^{\uparrow}$ & $ACC_{\rm FA}$$^{\uparrow}$\\
\midrule
    \textsc{GraspGPT}~\cite{tang2023graspgpt} & $0.72$ & $ - $ & $ - $ & $0.67$ 
                  & $0.56$ & $ - $ & $ - $ & $0.34$ \\
  \rowcolor{blue!10}
    \textsc{Ours} & $\mathbf{0.80}$ & $\mathbf{0.85}$ & $\mathbf{0.63}$ & $\mathbf{0.72}$ 
                  & $\mathbf{0.71}$ & $\mathbf{0.52}$ & $\mathbf{0.50}$ & $\mathbf{0.54}$ \\
\midrule
  \end{tabular}}
\end{table}

\begin{figure}[htbp]
  \centering
\includegraphics[width=1.0\linewidth]{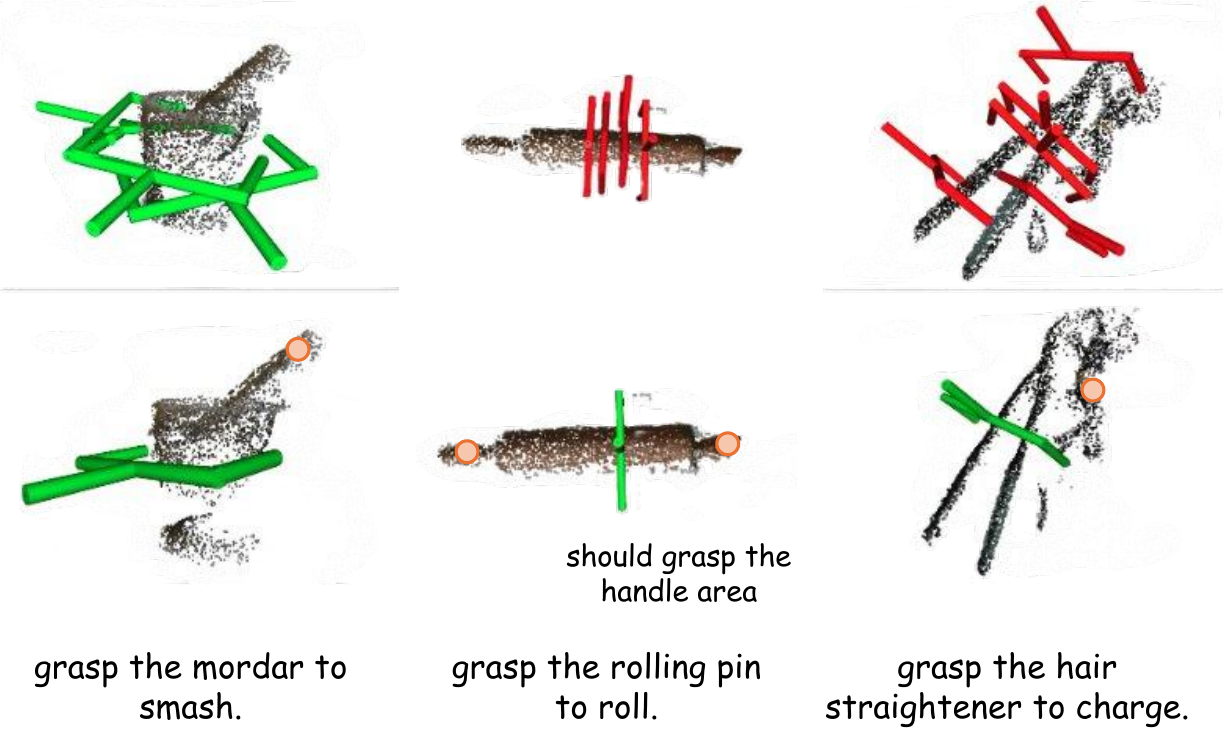}
  \caption{Inference results of GraspGPT~\cite{tang2023graspgpt}. The inference grasp poses are aligned based on the dexterous hand finger configuration for grasp validation.}\label{fig:inference_result_graspgpt}
\end{figure}

As illustrated in the inference examples (Fig.~\ref{fig:inference_result_graspgpt}), BERT-based GraspGPT~\cite{tang2023graspgpt} fails to identify appropriate grasp candidates within the designated regions when presented with complex semantic instructions.

\subsubsection{Real-World Experiments}
We further evaluate our OmniDexGraspNet in real world platform and compare with state-of-the-art methods, including GraspGPT~\cite{tang2023graspgpt}. The grasp generation results and corresponding semantic distribution are shown in Fig.~\ref{fig:result_grasp_taxonomy_real_world} and the quantitative results are summarized in Tab.~\ref{tab:results_simulation_comparison_grasping_exp}.

OmniDexGraspNet enables grasp generation guided by diverse semantic inputs, including grasp type, functional affordance, contact semantics, and finger configurations. However, in real-world experiments, we observe a degradation in contact accuracy and grasp taxonomy semantic accuracy. This performance drop is primarily attributed to the dynamic nature of contact during physical execution, where finger-object interactions often evolve over time. Consequently, such changes can alter the grasp type originally intended by the semantic prompt.

We also conduct experiments across different numbers of fingers and measured the corresponding grasp success rates, as shown in Tab.~\ref{tab:exp_different_finger_configuration}. The results show a clear trend: grasp success rate increases progressively with the number of fingers.

\begin{table}[t]
\centering
\caption{Grasp success rate across different finger configurations.}
\label{tab:exp_different_finger_configuration}
\begin{tabular}{lcccc}
\toprule
\textbf{Method} & \textbf{2-Finger} & \textbf{3-Finger} & \textbf{4-Finger} & \textbf{Overall} \\
\midrule
\textbf{Ours} & 64.98\% & 76.23\% & 78.83\% & 73.34\% \\
\bottomrule
\end{tabular}
\end{table}

\subsubsection{Failure Examples}

\begin{figure}[htbp]
  \centering
\includegraphics[width=1.0\linewidth]{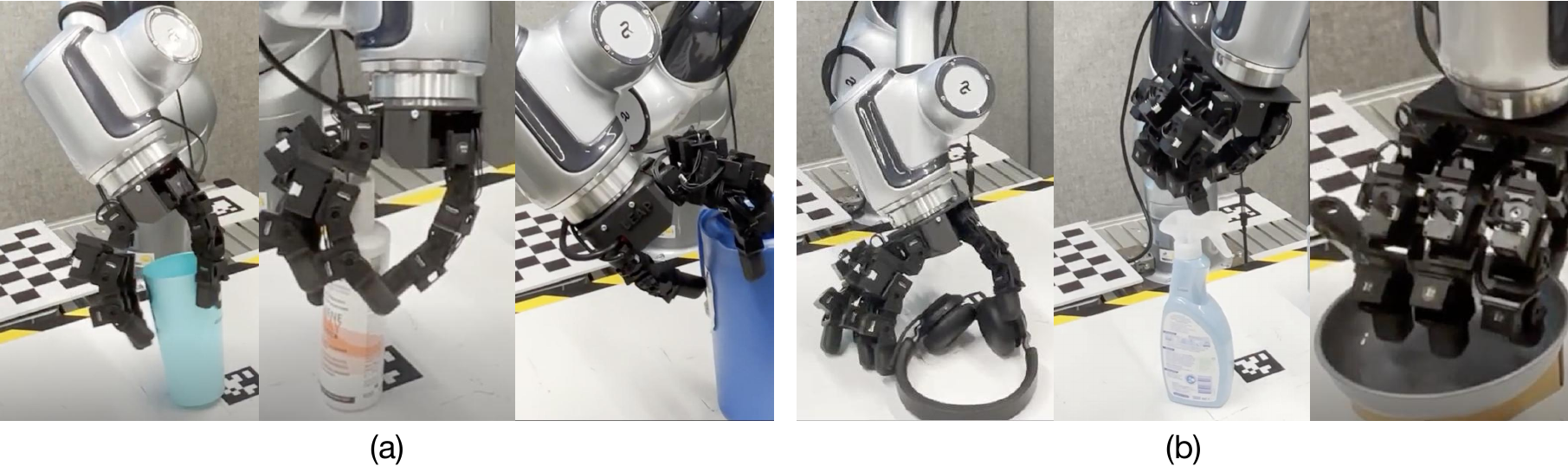}
  \caption{Example of failures. (a) Invalid force closure. (b) Object slippage.}\label{fig:inference_result_omnidexgraspnet}
\end{figure}

Failure cases are primarily attributed to suboptimal force closure in the generated grasps, or to object slippage during execution. As illustrated in the Fig.~\ref{fig:inference_result_omnidexgraspnet}, these failures often occur when the generated grasp lacks sufficient stability or fails to fully constrain the object, leading to unsuccessful grasp execution.

%%%%%%%%%%%%%%%%%%%%%%%%%%%%%%%%%%%%%%%%%%
\section{Conclusion and Future Works}
\label{sec:Conclude}
%%%%%%%%%%%%%%%%%%%%%%%%%%%%%%%%%%%%%%%%%%
In this work, we addressed the limited semantic modeling capability in existing dexterous grasp synthesis methods and introduced OmniDexVLG, a multimodal semantic-guided framework that explicitly incorporates grasp taxonomy, contact semantics, and functional affordance into the grasp generation process. 
This enables the synthesis of more semantically consistent and structurally diverse dexterous grasp poses under natural language instructions.

To support fine-grained semantic supervision, we proposed OmniDexDataGen, a comprehensive grasp data generation pipeline that integrates taxonomy-driven topological configuration sampling, functional-affordance-aware contact point sampling, taxonomy-aware differential force closure grasp sampling, and physics-based optimization. This framework produces dexterous grasp samples that cover a broad range of semantic categories and contact structures. 

Furthermore, we developed OmniDexReasoner, a dexterous grasp semantic inference module that leverages large multimodal model reasoning together with RAG and CoT mechanisms to decode latent grasp intentions embedded in language and task context, enabling automatic and reliable semantic annotation.

We conducted extensive ablation studies to evaluate the contribution of each proposed component. The functional-affordance-aware contact sampler effectively prevents semantically invalid grasps—particularly on large-scale objects—and enhances both the validity and diversity of functional affordance distributions. The taxonomy-aware DFC sampler enriches grasp diversity across semantic dimensions by encouraging varied grasp types and contact configurations. Within OmniDexReasoner, RAG improves contextual grounding across multi-agent reasoning, while CoT enhances discrimination among semantically similar grasp types. Together, these mechanisms substantially improve grasp taxonomy inference accuracy and stability.

Comprehensive simulation and real-world experiments demonstrate that OmniDexVLG outperforms existing approaches in terms of grasp diversity, contact distribution quality, semantic consistency, and generalization across functional tasks, validating the effectiveness of the proposed multi-semantic modeling paradigm for dexterous grasp generation.

In future work, we plan to extend our framework to support task-chained language instructions and continuous manipulation trajectories, enabling more complex and semantically grounded robotic manipulation capabilities.

%%%%%%%%%%%%%%%%%%%%%%%%%%%%%%%%%%%%%%%%%%
\bibliography{main}
%%%%%%%%%%%%%%%%%%%%%%%%%%%%%%%%%%%%%%%%%%

%%%%%%%%%%%%%%%%%%%%%%%%%%%%%%%%%%%%%%%%%%
% \newpage

 

% \end{IEEEbiography}

\empty
% \newpage    % for script modification test
% formal format
% \input{biography}
% \input{biography_anoymous}

%%%%%%%%%%%%%%%%%%%%%%%%%%%%%%%%%%%%%%%%%%
\vfill
%%%%%%%%%%%%%%%%%%%%%%%%%%%%%%%%%%%%%%%%%%
%%%%%%%%%%%%%%%%%%%%%%%%%%%%%%%%%%%%%%%%%%
% \newpage
% \appendix
% \input{07_appendix}
%%%%%%%%%%%%%%%%%%%%%%%%%%%%%%%%%%%%%%%%%%
\end{document}